\documentclass[twoside,11pt]{article}

\usepackage{jmlr2e}
\usepackage{algorithm}
\usepackage{algorithmic}

\ShortHeadings{Generative Iterative Closest Points}{Rajam\"aki and H\"am\"al\"ainen}
\firstpageno{1}

\usepackage{color}
\usepackage{amsmath}
\usepackage{float}
\usepackage{graphicx}
\usepackage{subcaption}
\usepackage{placeins}
\usepackage{comment}

\newcommand{\bx}{\mathbf{x}}
\newcommand{\by}{\mathbf{y}}

\newcommand{\bz}{\mathbf{z}}
\newcommand{\byhat}{\widehat{\mathbf{y}}}

\newcommand{\rulesep}{\unskip\ \vrule\ }

\newcommand{\bp}{\mathbf{p}}
\newcommand{\bq}{\mathbf{q}}

\definecolor{cambridgeblue}{rgb}{0.64, 0.76, 0.68}
\definecolor{arylideyellow}{rgb}{0.91, 0.84, 0.42}
\definecolor{ao_english}{rgb}{0.0, 0.5, 0.0}
\definecolor{cadmiumorange}{rgb}{0.93, 0.53, 0.18}

\begin{document}

\title{An Iterative Closest Points Approach to Neural Generative Models}

\author{\name Joose Rajam\"aki \email joose.rajamaki@aalto.fi, joose@joose.info \\
       \addr Department of Computer Science\\
       Aalto University\\
       Helsinki, Finland
       \AND
       \name Perttu H\"am\"al\"ainen \email perttu.hamalainen@aalto.fi \\
       \addr Department of Computer Science\\
       Aalto University\\
       Helsinki, Finland}

\editor{}

\maketitle

\begin{abstract} 
We present a simple way to learn a transformation that maps samples of one distribution to the samples of another distribution. Our algorithm comprises an iteration of 1) drawing samples from some simple distribution and transforming them using a neural network, 2) determining pairwise correspondences between the transformed samples and training data (or a minibatch), and 3) optimizing the weights of the neural network being trained to minimize the distances between the corresponding vectors. This can be considered as a variant of the Iterative Closest Points (ICP) algorithm, common in geometric computer vision, although ICP typically operates on sensor point clouds and linear transforms instead of random sample sets and neural nonlinear transforms. We demonstrate the algorithm on simple synthetic data and MNIST data. We furthermore demonstrate that the algorithm is capable of handling distributions with both continuous and discrete variables.
\end{abstract} 

\begin{figure*}[h]
        \centering
        \begin{subfigure}{0.45\linewidth}
                \centering
                \includegraphics[width=\textwidth]{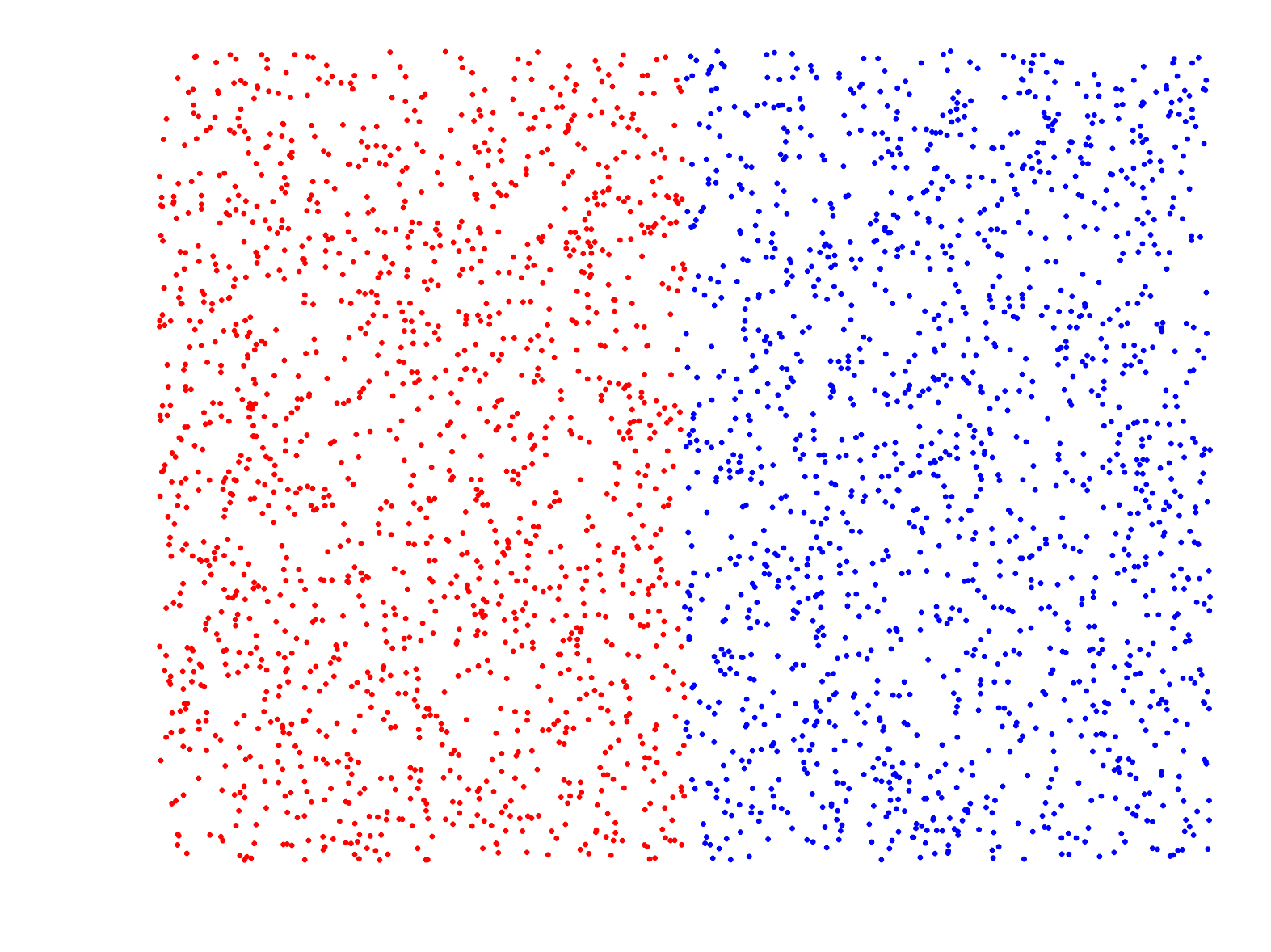}
                \caption{Samples from the origin distribution.}
        \end{subfigure}%
        {\LARGE$\Rightarrow$}
        \begin{subfigure}{0.45\linewidth}
                \centering
                \includegraphics[width=\textwidth]{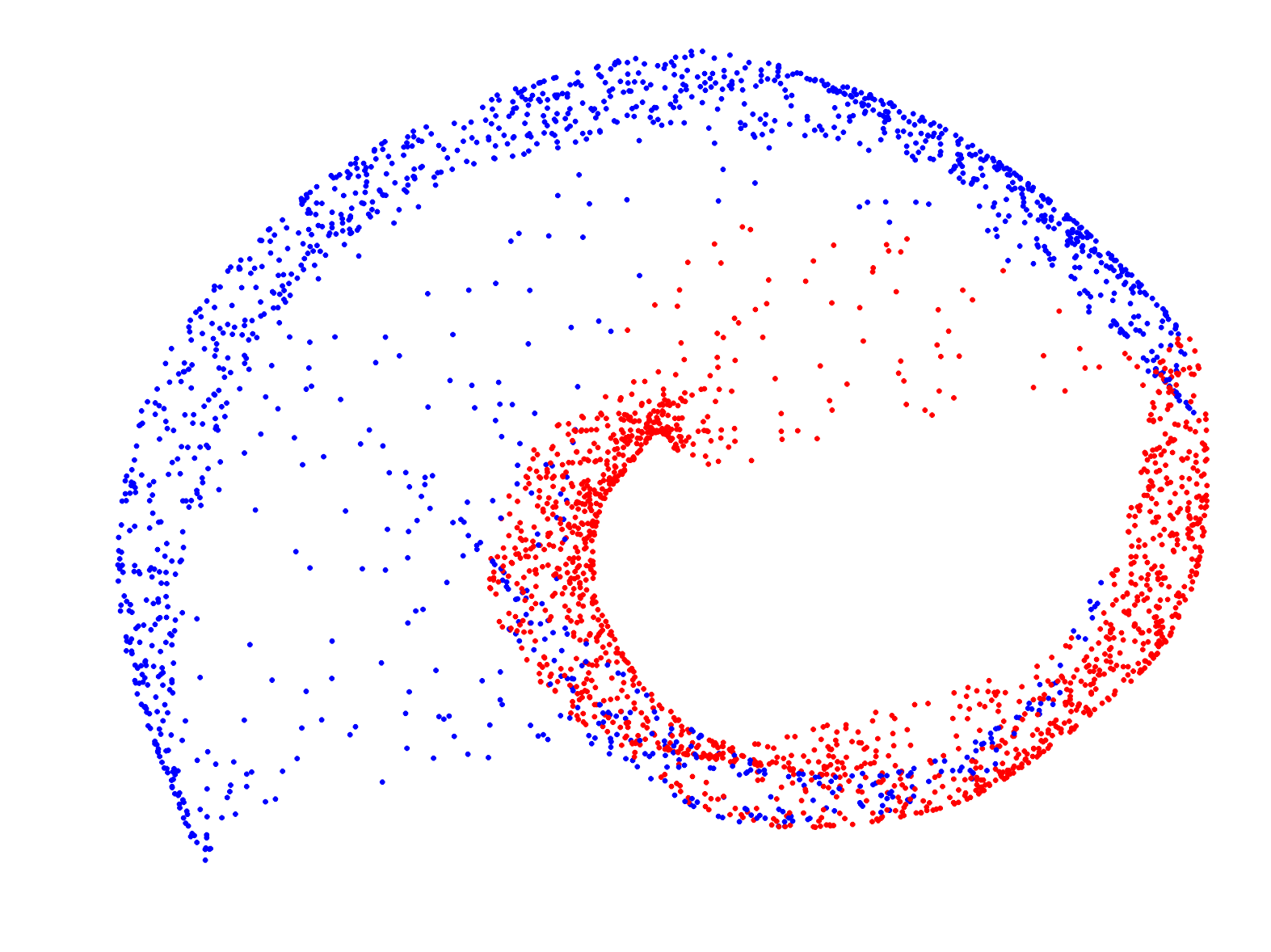}
                \caption{Generated samples.}
        \end{subfigure}%
        ~ 
        \caption{A two dimensional uniform distribution is transformed to a swiss roll with added noise. \label{fig:teaser}}
\end{figure*}

\section{Introduction}
\label{sec:intro}

Learning transformations between distributions has recently been studied extensively because it allows drawing samples from a simple distribution and mapping them to a more complex distribution of interest. To train neural networks to map distributions to other distributions is usually performed by generative adversarial network training (GAN) \citep{Goodfellow2014}. GANs are, however, notoriously difficult to train because of the unstable dynamics between two competing networks.

We present a simple and stable algorithm to train transformations between distributions such as the one demonstrated in Figure \ref{fig:teaser}. In contrast to GANs, our training method involves just one network. The contribution of this paper is the training algorithm that yields robust training of neural networks for mapping between distributions. Our algorithm comprises an iteration of 1) drawing samples from some simple distribution and transforming them using a neural network, 2) determining pairwise correspondences between the transformed samples and training data (or a minibatch), and 3) optimizing the weights of the neural network being trained to minimize the distances between the corresponding vectors. This can be considered as a variant of the Iterative Closest Points (ICP) algorithm, common in geometric computer vision, although ICP typically operates on sensor point clouds and linear transforms instead of random sample sets and neural nonlinear transforms.

Our algorithm works well also with conditioned data generation and we can also use it to map mixed distributions to discrete ones. To the best of our knowledge, this is the first algorithm to demonstrate mappings from a mixed distribution to a discrete distribution using just one neural network.

An implementation of the algorithm is available in Github: \linebreak https://github.com/JooseRajamaeki/ICP. 


\section{Related methods}
\label{sec:related_methods}

One of the most succesful approaches in generative models are generative adversarial networks \citep{Goodfellow2014}. Generative adversarial training of neural networks involves training a generator network to map some easy-to-sample distribution to another distribution. The training of GANs also involves training a discriminator network that tries to tell apart the data from the target distribution and the data generated by the generator network. This idea is simple and brilliant but the training involves of technical challenges. The most notable challenge is coordinating the speed at which each of the networks is trained. It happens easily that the discriminator network quickly learns to tell apart the true data and the generated data after which it does not provide a strong gradient to the generator network. These problems have been alleviated to some extent by Wasserstein-GANs \citep{Arjovsky2017} and their extensions \citep{Gulrajani2017}. We do not have any problems arising from the unstable dynamics of two networks. Furthermore, since we are minimizing a distance measure, we have a strong gradient in all sensible cases.

Optimizing latent variable space can be performed to obtain fast to train algorithm with results that are competitive with GANs \citep{Bojanowski2017}. The algorithm trains only one neural network, like we do. The algorithm learns the latent space, though, i.e. one is not free to choose the sampling distribution but it is instead intractable and given by the training algorithm.

Generative moment matching networks (GMMN) \citep{Li2015generative} are a successful approach to generative models. They also involve training only one neural network, which uses the objective of matching the moments of the generated data and those of the true data. Like our algorithm, the algorithm necessitates the existence of a meaningful distance in the space of the generated data. However, GMMNs involve additionally choosing a kernel width parameter, which requires prior insight to the structure of the  data.

Using bijective mappings has been used to build models for density estimation and sampling \citep{Dinh2014,Dinh2016}. They map a distribution to a latent space with a simple structure. Density estimation in the latent space is straightforward and sampling can be performed effectively as ancestral sampling. Our training algorithm does not impose any restrictions on the model being trained. For instance we can train neural networks to map distributions from continuous to discrete, for which one cannot train bijective mappings.

Variational autoencoders (VAE) \citep{Kingma2013} are another successful approach to train generative models. They utilize the reparametrization trick to train an encoder and a decoder by backpropagation. The encoder transforms the data points to the latent space and the decoder transforms the latent space variables back to the data space. In contrast to our algorithm, the variational autoencoders also involve training two networks. Furthermore, the reparametrization trick sets restrictions to the distributions involved, such as demands for continuity. There are recent advances in extending the variational autoencoders to discrete distributions \citep{Jang2017,Maddison2016}. This involves further approximations, though, whereas our algorithm extends naturally to discrete distributions as we demonstrate in Section \ref{sec:continuous_to_discrete}. Our algorithm offers near complete freedom to choose the distributions involved and the associated model. In Section \ref{sec:results} we use a mixed distribution with discrete and continuous variables as the input "noise" distribution.

The methods described above involve training a network to map from one distribution to other. Also other generative models and ways to use neural networks exist. For example \citep{Bordes2017,Goyal2017variational} present iterative denoising approaches to sample generation. \citep{Bordes2017} also offer a comprehensive review of the techniques used in generative modeling.

Our method is reminiscent of the Iterative Closest Points (ICP) algorithm \citep{Chen1991,Besl1992}, common in computer vision geometry processing. ICP alternates between 1) computing pairwise correspondences between two sets of vectors, and 2) optimizing a transform that minimizes the distance between corresponding vectors. In ICP, the vector sets are typically point clouds from some sensor(s), whereas in our case, one set is a batch of training data, and the other set is drawn from some simple distribution and transformed by a neural network to the training data space. In our case, the sets are also drawn independently for each iteration whereas they stay the same in each ICP iteration. Furthermore, in ICP the optimized transform is usually a simple geometric one, for which a closed-form solution exists. In contrast to that, we optimize the neural network weights through backpropagation with the sum of pairwise distances as the loss.


\section{Methods}
\label{sec:methods}

This section presents our algorithms. As stated before we have developed an algorithm for learning a mapping to transform samples drawn from one distribution to another distribution. We can also learn a mapping conditioned on a subset of the training data variables.

\subsection{Notational conventions and the problem setting}
\label{sec:notational_conventions}

Noise from origin or "noise" distribution $\mathcal{D}_\mathrm{origin}$ is denoted by $\mathbf{x}$, and $\by$ are the values from the target distribution $\mathcal{D}_\mathrm{target}$, i.e. the training data. The conditioning part of the target distribution $\mathcal{D}_\mathrm{target}$ is denoted by $\mathbf{z}$, if such values are needed. For the unconditioned algorithm, the goal is to find such a mapping $f$ that:
\begin{equation}
\mathbf{y} = f \left( \mathbf{x} \right), \; \mathbf{y} \sim \mathcal{D}_\mathrm{target}, \; \mathbf{x} \sim \mathcal{D}_\mathrm{origin}.
\end{equation}
For the conditioned algorithm we are training $f$ such that:
\begin{equation}
\begin{bmatrix}
\mathbf{z}\\
\mathbf{y}
\end{bmatrix} =
f \left( 
\begin{bmatrix}
\mathbf{z}\\
\mathbf{x}
\end{bmatrix}
\right), \;
\begin{bmatrix}
\mathbf{z}\\
\mathbf{y}
\end{bmatrix} \sim \mathcal{D}_\mathrm{target}, \;
\mathbf{x} \sim \mathcal{D}_\mathrm{origin}.
\end{equation}

Our algorithm necessitates a meaningful distance $d(\by_1,\by_2)$ for the space in which the data points $\by_1$ and $\by_2$ reside. We furthermore assume that the trainable function approximator $f$ is continuous. In our tests the function approximator was chosen to be a neural net. We use the circumflex to denote the variables that we get by mapping some variables with $f$, for example $\widehat{y} = f(x)$.

Our aim is to build an algorithm that works in the same manner as GANs. Given samples $\{ \by_0, \by_1, ... , \by_N \}$ we wish to train the mapping $f$ such that it produces "predictions" $\widehat{\by}$ that are indistinguishable from the given samples. We assume that the probability density function or the probability mass function of the distribution being approximated is not known. When it is known, efficient algorithms to train $f$ exist \citep{Liu2016}.

\subsection{The iterative closest points for distribution training algorithm}
\label{sec:output_matching_algorithm}

\begin{figure*}[h]
        \centering
        \begin{subfigure}[t]{0.23\linewidth}
                \centering
                \includegraphics[width=\textwidth]{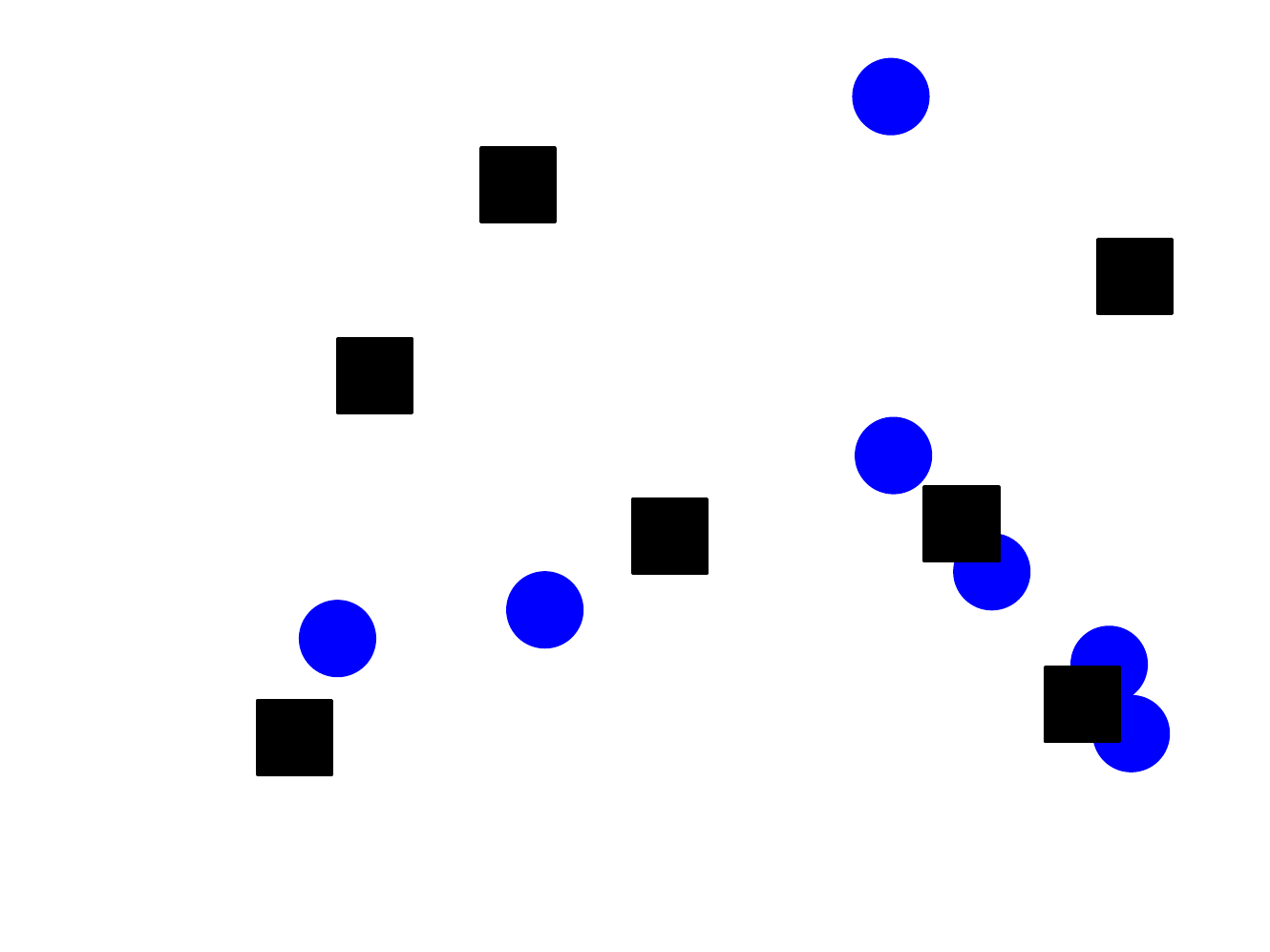}
                \caption{Initial situation. \textcolor{blue}{Blue} denotes the training data and black denotes the generated predictions.}
                \label{fig:illustration_init}
        \end{subfigure}%
        \rulesep
        \begin{subfigure}[t]{0.23\linewidth}
                \centering
                \includegraphics[width=\textwidth]{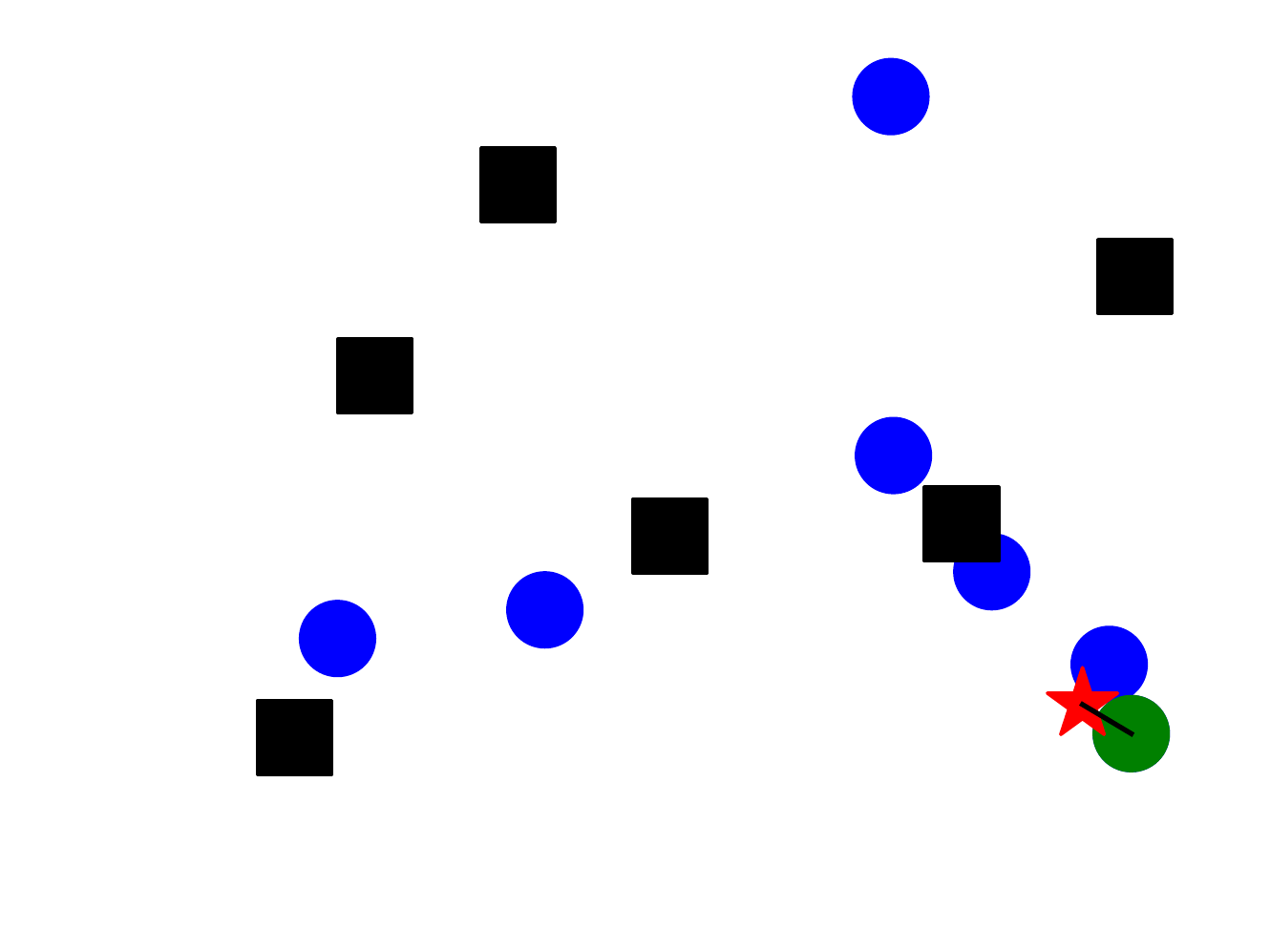}
                \caption{The first step. We choose a random data point $\by$ (\textcolor{ao_english}{green} dot) and find the closest prediction $\widehat{\by}$ (\textcolor{red}{red} asterisk). These form a pair drawn as a black line.}
                \label{fig:illustration_first_step}
        \end{subfigure}%
        \rulesep
        \begin{subfigure}[t]{0.23\linewidth}
                \centering
                \includegraphics[width=\textwidth]{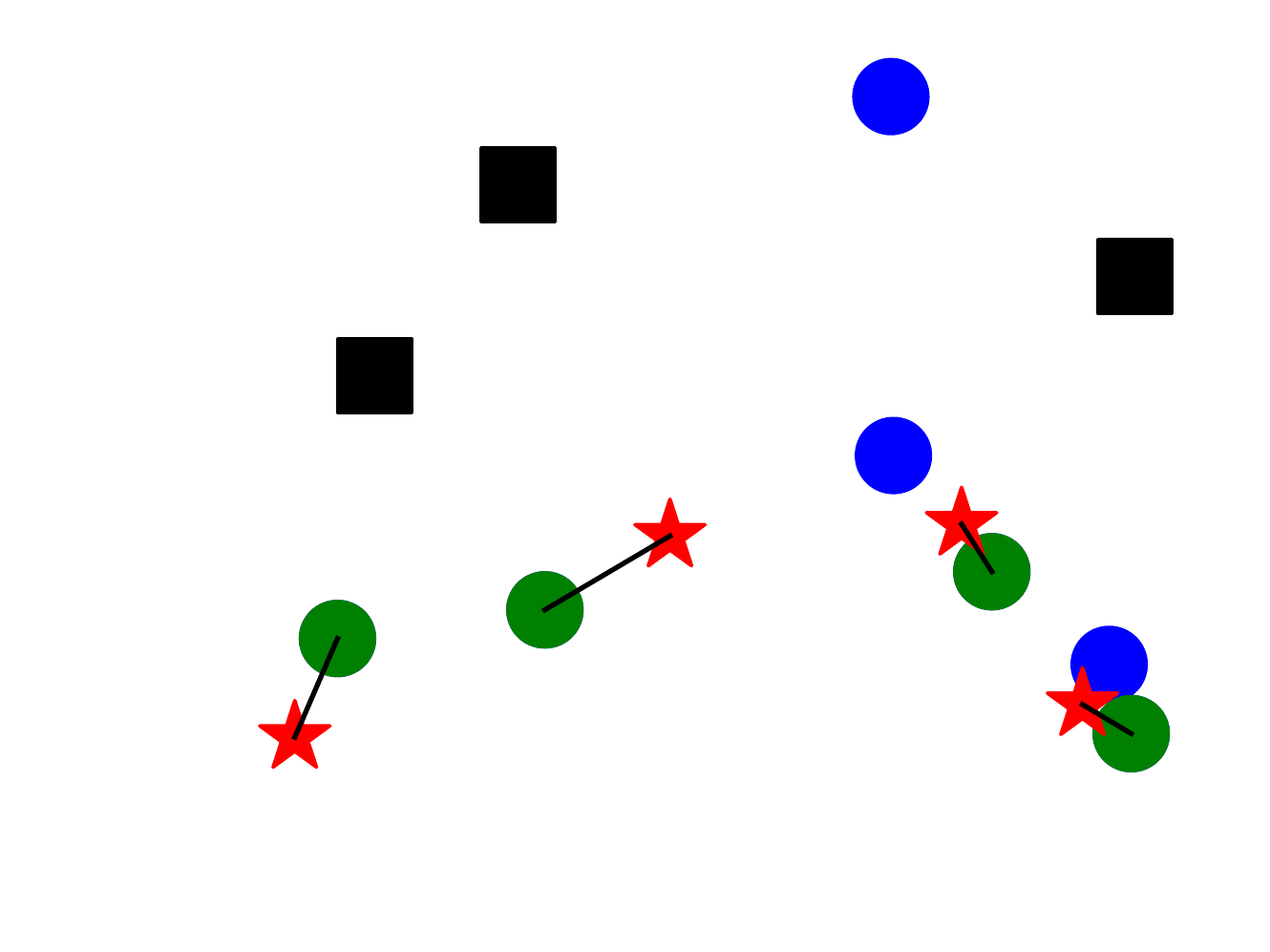}
                \caption{The point matching procedure continues.}
                \label{fig:illustration_fourth_step}
        \end{subfigure}%
        \rulesep
        \begin{subfigure}[t]{0.23\linewidth}
                \centering
                \includegraphics[width=\textwidth]{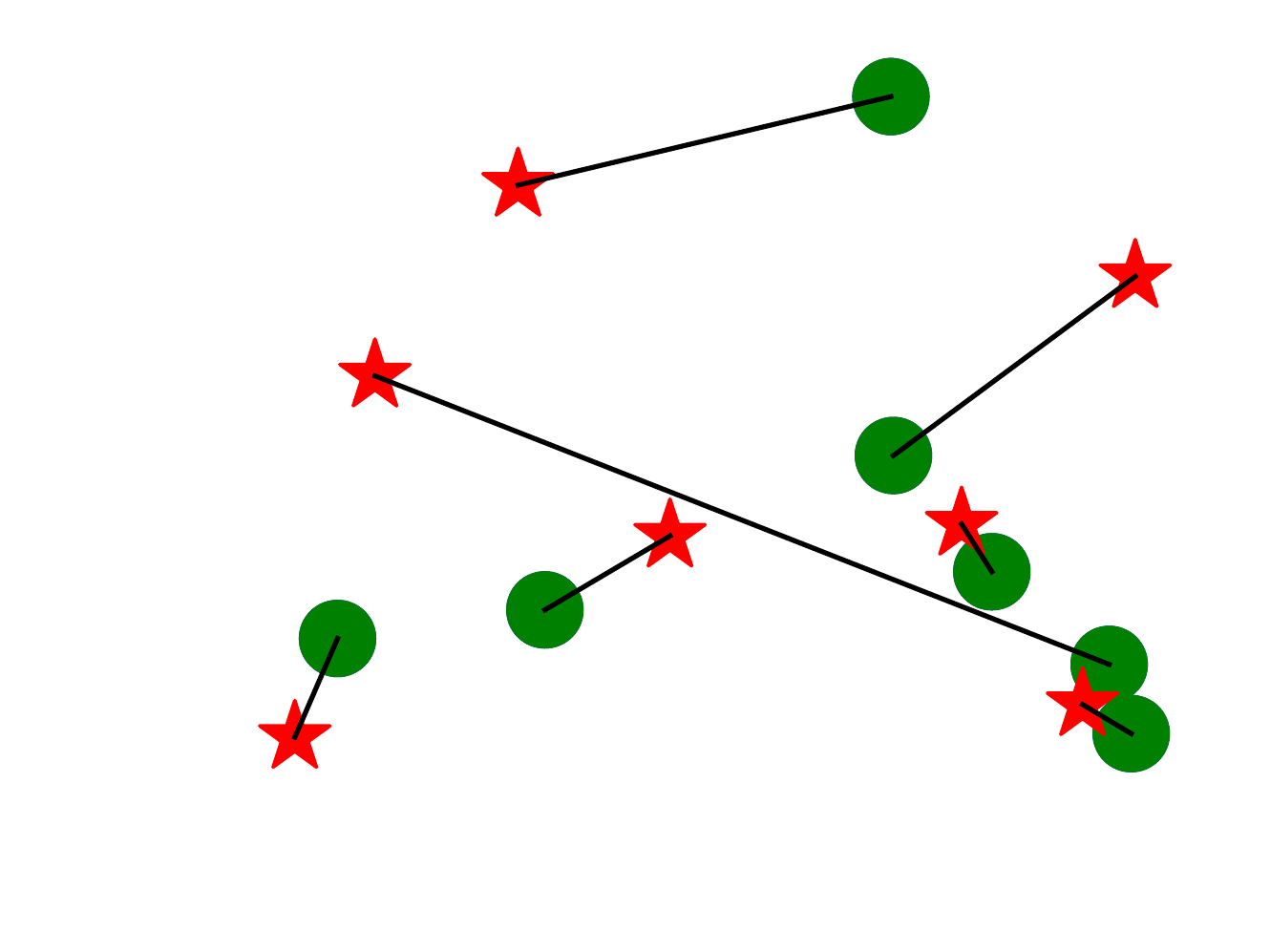}
                \caption{The point matching is complete. The last data point gets matched to the last available prediction. Note that the last data points may get matched to predictions that are quite far.}
                \label{fig:illustration_last_step}
        \end{subfigure}%
        
        ~ 
        \caption{An illustrative example of the idea of the algorithm. The predictions $\widehat{\by}$ (black squares) are matched to the actual data (\textcolor{blue}{blue} dots). \label{fig:learning_illustration}}
\end{figure*}

Our method is described as pseudocode in Algorithm \ref{alg:incremental_matching}. Its conditioned version is presented in Algorithm \ref{alg:conditioned_incremental_matching}. The algorithms work by choosing random data point from the true data and finding the closest mapped point. This is performed incrementally such that the mapped points that have already been matched to some data point are not considered in the matching procedure. When all the data points have been matched, the mapping $f$ is trained with supervised learning. Figure \ref{fig:learning_illustration} illustrates the algorithm.

\begin{algorithm}
   \caption{The iterative closest points for distribution training algorithm}
   \label{alg:incremental_matching}
\begin{algorithmic}[1]
   \REPEAT
   
   \STATE Sample set $\mathcal{D}_\mathrm{batch\_origin} = \{\bx_0,\bx_1,...,\bx_N\}$ from $\mathcal{D}_\mathrm{origin}$
   \STATE Sample set $\mathcal{D}_\mathrm{batch\_target} = \{\by_0,\by_1,...,\by_N\}$ from $\mathcal{D}_\mathrm{target}$
   
   \STATE Initialize $\mathcal{D}_\mathrm{pairs} = \emptyset$
   
   \FORALL{$\bx_i \in \mathcal{D}_\mathrm{batch\_origin}$}
		\STATE Perform the mapping $\byhat_i = f(\bx_i)$
		\STATE Add pair $(\bx_i,\byhat_i)$ to $\mathcal{D}_\mathrm{pairs}$
   \ENDFOR
   
   \STATE Initialize $\mathcal{D}_\mathrm{ordered} = \emptyset$
   \FORALL{$\by_i \in \mathcal{D}_\mathrm{batch\_target}$}
   		\STATE Find index $j = \arg \min d( \by_i , \byhat_j ) $
   		\STATE Add pair $(\bx_j,\by_i)$ to $\mathcal{D}_\mathrm{ordered}$
   		\STATE Remove pair $(\bx_j,\byhat_j)$ from $\mathcal{D}_\mathrm{pairs}$
   \ENDFOR

   \STATE Train $f$ with supervised learning using $\mathcal{D}_\mathrm{ordered}$
   
   \UNTIL{Convergence}
\end{algorithmic}
\end{algorithm}
\begin{algorithm}
   \caption{The conditioned iterative closest points for distribution training algorithm}
   \label{alg:conditioned_incremental_matching}
\begin{algorithmic}[1]
   \REPEAT
   
   \STATE Sample set $\mathcal{D}_\mathrm{batch\_origin} = \{\bx_0,\bx_1,...,\bx_N\}$ from $\mathcal{D}_\mathrm{origin}$
   \STATE Sample set \\$\mathcal{D}_\mathrm{batch\_target} = \left\{\begin{bmatrix}\bz_0 \\ \by_0\end{bmatrix},\begin{bmatrix}\bz_1 \\ \by_1\end{bmatrix},...,\begin{bmatrix}\bz_N \\ \by_N\end{bmatrix} \right\}$ \\from $\mathcal{D}_\mathrm{target}$
   \STATE Initialize $\mathcal{D}_\mathrm{pairs} = \emptyset$
   
   \FOR{$i = 0 ... N$}
		\STATE Perform the mapping $\begin{bmatrix}\widehat{\bz}_i \\ \byhat_i\end{bmatrix} = f\left( \begin{bmatrix}\bz_i \\ \bx_i \end{bmatrix} \right)$
		\STATE Add pair $\left(\begin{bmatrix}\bz_i \\ \bx_i \end{bmatrix},\begin{bmatrix}\bz_i \\ \byhat_i\end{bmatrix} \right)$ to $\mathcal{D}_\mathrm{pairs}$
   \ENDFOR
   
   \STATE Initialize $\mathcal{D}_\mathrm{ordered} = \emptyset$
   \FORALL{$\by_i \in \mathcal{D}_\mathrm{batch\_target}$}
   		\STATE Find index $j = \arg \min d \left( \begin{bmatrix}\bz_i \\ \by_i\end{bmatrix}, \begin{bmatrix}\bz_j \\ \byhat_j\end{bmatrix} \right) $
   		\STATE Add pair $\left(\begin{bmatrix}\bz_i \\ \bx_j \end{bmatrix},\begin{bmatrix}\bz_i \\ \by_i\end{bmatrix} \right)$ to $\mathcal{D}_\mathrm{ordered}$
   		\STATE Remove pair $\left(\begin{bmatrix}\bz_j \\ \bx_j \end{bmatrix},\begin{bmatrix}\bz_j \\ \byhat_j\end{bmatrix} \right)$ from $\mathcal{D}_\mathrm{pairs}$
   \ENDFOR

   \STATE Train $f$ with supervised learning using $\mathcal{D}_\mathrm{ordered}$
   
   \UNTIL{Convergence}
\end{algorithmic}
\end{algorithm}

\begin{figure*}[h]
        \centering
        \begin{subfigure}{0.23\linewidth}
                \centering
                \includegraphics[width=\textwidth]{image_iteration_0.pdf}
                \caption{Initial situation.}
        \end{subfigure}%
        \rulesep
        \begin{subfigure}{0.23\linewidth}
                \centering
                \includegraphics[width=\textwidth]{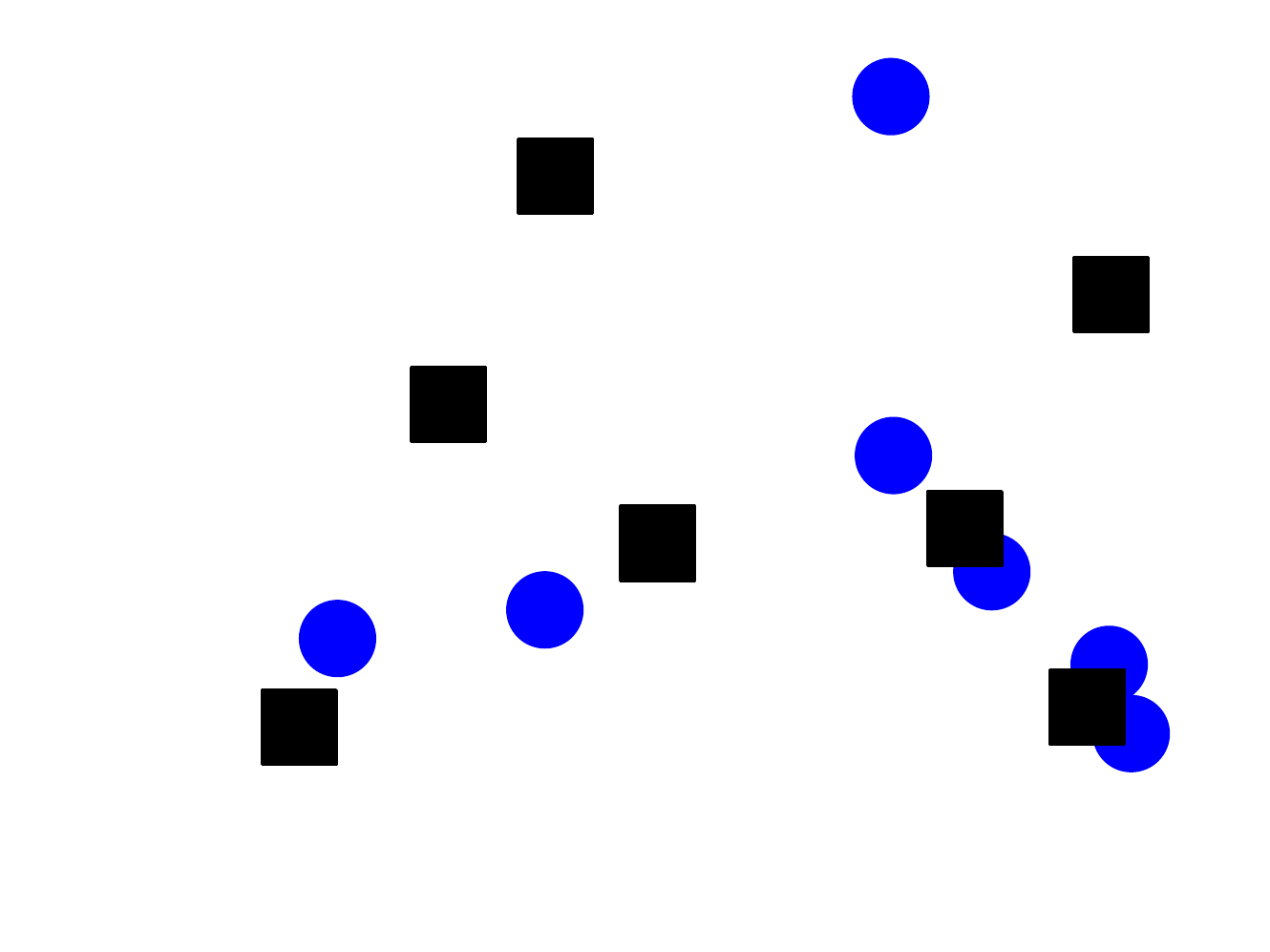}
                \caption{First epoch.}
        \end{subfigure}%
        \rulesep
        \begin{subfigure}{0.23\linewidth}
                \centering
                \includegraphics[width=\textwidth]{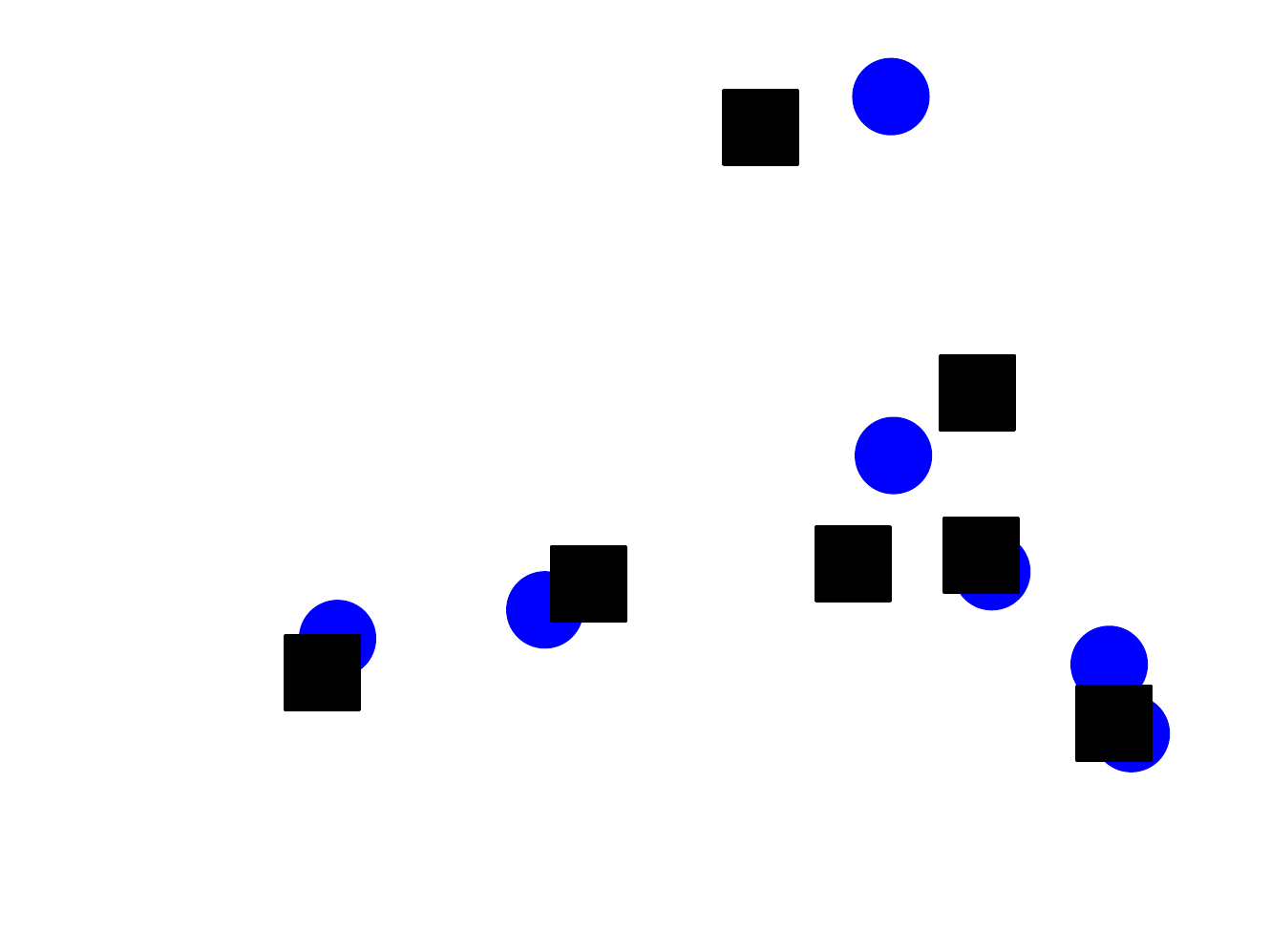}
                \caption{Epoch 10.}
        \end{subfigure}%
        \rulesep
        \begin{subfigure}{0.23\linewidth}
                \centering
                \includegraphics[width=\textwidth]{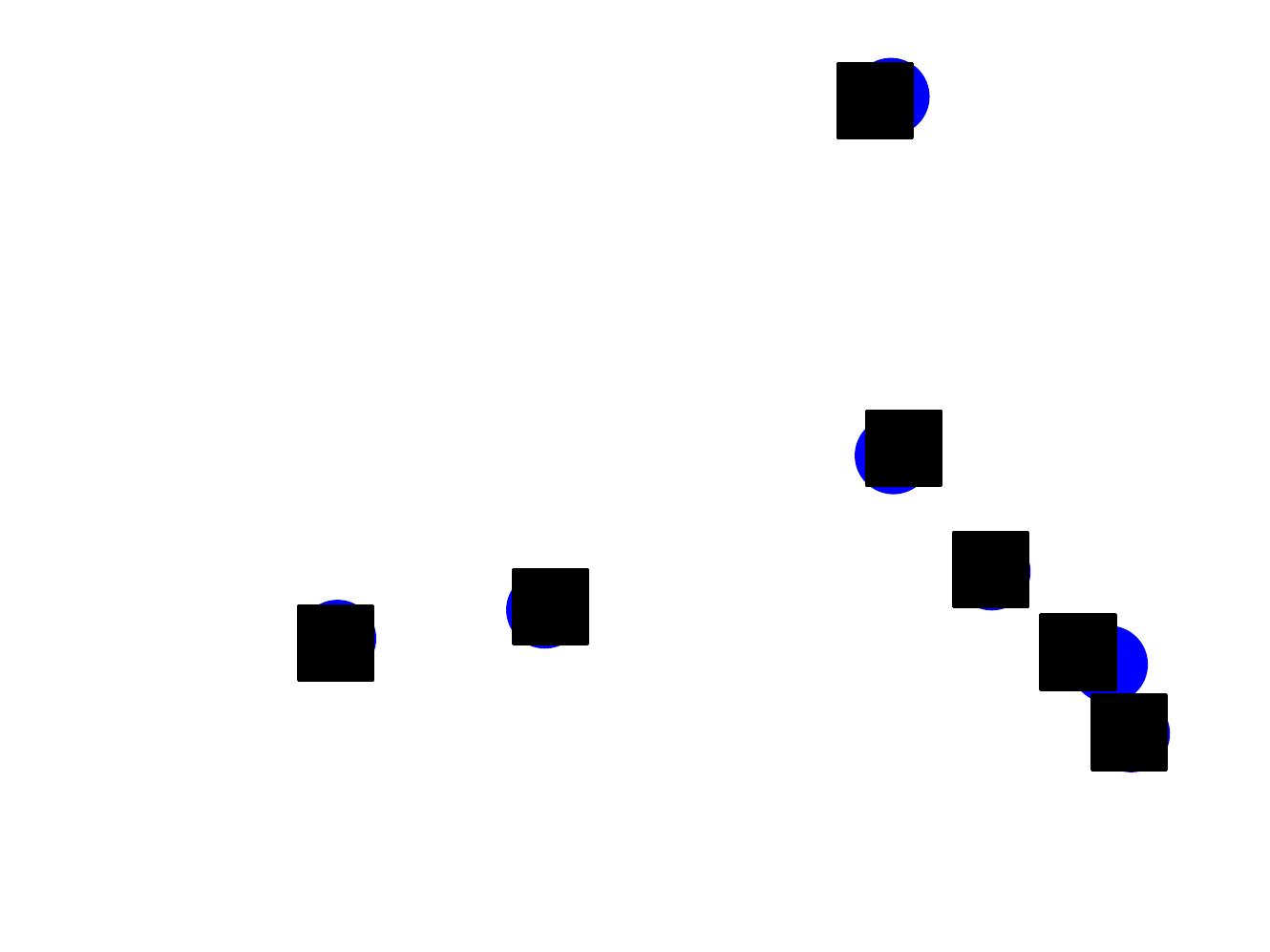}
                \caption{Epoch 30.}
        \end{subfigure}%
        ~ 
        \caption{An illustrative example of how the predictions $\widehat{\by}$ (black squares) will start to get closer to a point of actual data (\textcolor{blue}{blue} dots). \label{fig:evolution}}
\end{figure*}

The last points to be matched in Algorithms \ref{alg:incremental_matching} and \ref{alg:conditioned_incremental_matching} get poorly matched with a very high probability. However, even if some points get matched poorly, the well-matched points in their surrounding will draw them to the right directions. Because the mapping $f$ is continuous, similar points in the input space map to similar points in the output space, i.e. $f(\bx_i) \approx f(\bx_i + \boldsymbol{\varepsilon})$. This implies that $f$ can be thought of as a complicated weighting function of a self-organizing map \citep{Kohonen1982}.

\subsection{Convergence}
\label{sec:convergence}

The algorithm produces a matching of the mapped points $\byhat$ and the true data points $\by$ such that the matched points can be used to form an empirical estimate of the upper bound for the earth mover distance (EMD), i.e. the Wasserstein metric. Given such a matching, the upper bound for the earth mover distance can be minimized by supervised learning. The earth mover distance between distributions $\mathcal{P}$ and $\mathcal{Q}$ is:
\begin{equation}
\displaystyle \text{EMD}(\mathcal{P},\mathcal{Q}) = \min_{w(\bp,\bq)} \frac{\displaystyle \int_{\bp \in \mathcal{P}} \int_{\bq \in \mathcal{Q}} w(\bp,\bq) d(\bp,\bq)}{\displaystyle \int_{\bp \in \mathcal{P}} \int_{\bq \in \mathcal{Q}} w(\bp,\bq)}.
\end{equation}
Here $w$ is the function denoting the flow, i.e. $w(\bp,\bq)$ is the amount of (probability) mass that needs to be transferred from $\bq$ to $\bp$. The distance from $\bq$ to $\bp$ is denoted by $d(\bp,\bq)$.

Suppose that we have two equally big empirical samples from the two distributions at hand:
\begin{eqnarray}
\{ \bp_0, \bp_1, ..., \bp_N \} &\sim & \mathcal{P} \label{eq:sample_p}\\
\{ \bq_0, \bq_1, ..., \bq_N \} &\sim & \mathcal{Q}. \label{eq:sample_q}
\end{eqnarray}
The earth mover distance for the distributions $\mathcal{P}$ and $\mathcal{Q}$ based on these empirical samples would be:
\begin{eqnarray}
\text{EMD}_\text{empirical}(\mathcal{P},\mathcal{Q}) =  \min \frac{\sum_{i=0}^{N}\sum_{j=0}^{N} \widetilde{w}(\bp_i,\bq_j) d(\bp_i,\bq_j)}{N+1}. \label{eq:empirical_emd}
\end{eqnarray}
Here $\widetilde{w}(\bp_i,\bq_j)$ is an indicator function, which outputs one when $\bq_j$ should be transferred to $\bp_i$ and zero otherwise. The indicator function $\widetilde{w}(\bp_i,\bq_j)$ tells the optimal transport scheme to transfer the points sampled from distribution $\mathcal{P}$ to the points sampled from distribution $\mathcal{Q}$. The indicator function $\widetilde{w}(\bp_i,\bq_j)$ satisfies the following conditions:
\begin{eqnarray}
\widetilde{w}(\bp_i,\bq_j) = 1 \Rightarrow \widetilde{w}(\bp_k,\bq_j) = 0 \; \text{if}\; i \neq k \label{eq:matching_condition_1}\\
\widetilde{w}(\bp_i,\bq_j) = 1 \Rightarrow \widetilde{w}(\bp_i,\bq_k) = 0 \; \text{if}\; j \neq k \label{eq:matching_condition_2}\\
\forall \bq_j \; \exists \bp_i \; \text{such that}\; \widetilde{w}(\bp_i,\bq_j) = 1\\
\forall \bp_i \; \exists \bq_j \; \text{such that}\; \widetilde{w}(\bp_i,\bq_j) = 1
\end{eqnarray}
This means that each data point from $\mathcal{Q}$ can only be matched to a single point from the distribution $\mathcal{P}$ and vice versa, and additionally each data point from both of the distributions needs to matched once. To phrase it briefly, the matching function $\widetilde{w}(\bp_i,\bq_j)$ is bijective.

Constructing the indicator function $\widetilde{w}(\bp_i,\bq_j)$ to compute the empirical estimate in Equation \eqref{eq:empirical_emd} is an assignment problem and it can be performed using the Hungarian algorithm \citep{Kuhn1955hungarian}.\footnote{The version of the algorithm where the matching is done by the Hungarian algorithm has been described by \cite{Bojanowski2017unsupervised}. They used it to learn the spherical uniform distribution. However, for them this was a proxy objective in unsupervised learning.} The Hungarian algorithm has the computational complexity $\mathcal{O}(N^3)$. However, we can compute estimates of the upper bound using {\bf any} bijective indicator function $\overline{w}(\bp_i,\bq_j)$. By definition:
\begin{eqnarray}
\sum_{i=0}^{N}\sum_{j=0}^{N} \widetilde{w}(\bp_i,\bq_j) d(\bp_i,\bq_j) \leq \sum_{i=0}^{N}\sum_{j=0}^{N} \overline{w}(\bp_i,\bq_j) d(\bp_i,\bq_j).
\end{eqnarray}
The Algorithms \ref{alg:incremental_matching} and \ref{alg:conditioned_incremental_matching} are based on constructing $\overline{w}(\by_i,\byhat_j)$, which is an assignment between the mapped points and the true data distribution. The assignment by matching the closest points of the remaining sets is $\mathcal{O}(N^2)$. After constructing $\overline{w}(\by_i,\byhat_j)$, the algorithms proceed to minimize the following loss function by supervised learning:
\begin{equation}
\sum_{i=0}^{N}\sum_{j=0}^{N} \overline{w}(\by_i,\byhat_j) d(\by_i,\byhat_j). \label{eq:loss_function}
\end{equation}
Thus, if we manage to learn a mapping $f$, which makes the loss function \eqref{eq:loss_function} zero, the predictions $\byhat$ follow exactly the target distribution $\mathcal{D}_\text{target}$. Intuitively this is a clear result; if each prediction $\byhat$ corresponds to one data point $\by$ sampled from the target distribution and is its exact replica, the distribution are the same. 

With a mapping $f$ that has unlimited capacity, we could use any bijective $\overline{w}(\by_i,\byhat_j)$, as stated earlier. However, continuous mappings such as most neural networks cannot be easily trained to approximate highly varying mappings that are likely to arise if one chooses $\overline{w}(\by_i,\byhat_j)$ arbitrarily, for example to be a random matching. Furthermore, a random matching is not very consistent between iterations, which makes the learning task harder. The closest points matching promotes the following properties, which make it easier to train a neural network to learn the task:
\begin{itemize}
\item points that are close to each other in the input space $\bx$ tend to be matched to points that are close to each other in the output space $\by$,
\item we try to avoid shifting the predictions $\byhat$ by large distances, and
\item $\overline{w}(\by_i,\byhat_j)$ should be consistent between iterations.
\end{itemize}

\subsection{Alternating versions of the matching algorithms}

Any bijective matching being possible inspired a modified version of the algorithm. Algorithms \ref{alg:incremental_matching} and \ref{alg:conditioned_incremental_matching} present the alternating versions of the closest points matching algorithms. The alternating versions switch between:
\begin{itemize}
\item picking a data point from the true data and finding the closest mapped point, and
\item picking a mapped point from and finding the closest true data point.
\end{itemize}
The alternating variant of the algorithm has the benefit that the same mapped data points will not be matched last over and over again.

\begin{algorithm}
   \caption{The alternating iterative closest points for distribution training algorithm}
   \label{alg:alternating_incremental_matching}
\begin{algorithmic}[1]
   \REPEAT
   
   \STATE Sample set $\mathcal{D}_\mathrm{batch\_origin} = \{\bx_0,\bx_1,...,\bx_N\}$ from $\mathcal{D}_\mathrm{origin}$
   \STATE Sample set $\mathcal{D}_\mathrm{batch\_target} = \{\by_0,\by_1,...,\by_N\}$ from $\mathcal{D}_\mathrm{target}$
   
   \STATE Initialize $\mathcal{D}_\mathrm{pairs} = \emptyset$
   
   \FORALL{$\bx_i \in \mathcal{D}_\mathrm{batch\_origin}$}
		\STATE Perform the mapping $\byhat_i = f(\bx_i)$
		\STATE Add pair $(\bx_i,\byhat_i)$ to $\mathcal{D}_\mathrm{pairs}$
   \ENDFOR
   
   \STATE Initialize $\mathcal{D}_\mathrm{ordered} = \emptyset$

   \WHILE{$|\mathcal{D}_\mathrm{batch\_target}| > 0$}
        \STATE Sample random number $b \sim \mathrm{Bernoulli}(0.5)$
        \IF { $b = 1$}
        	\STATE Pick random $\by_i \in \mathcal{D}_\mathrm{batch\_target}$
        	\STATE Find index $\displaystyle j = \arg \left[ \min_{(\bx_j,\byhat_j) \in \mathcal{D}_\mathrm{pairs}} d( \by_i , \byhat_j ) \right] $
        \ELSE
        	\STATE Pick random $(\bx_j,\byhat_j) \in \mathcal{D}_\mathrm{pairs}$
        	\STATE Find index $\displaystyle i = \arg \left[ \min_{\by_i \in \mathcal{D}_\mathrm{batch\_target}} d( \by_i , \byhat_j ) \right] $
        \ENDIF
        \STATE Add pair $(\bx_j,\by_i)$ to $\mathcal{D}_\mathrm{ordered}$
       	\STATE Remove pair $(\bx_j,\byhat_j)$ from $\mathcal{D}_\mathrm{pairs}$
   		\STATE Remove $\by_i$ from $\mathcal{D}_\mathrm{batch\_target}$
   \ENDWHILE   
   
   \STATE Train $f$ with supervised learning using $\mathcal{D}_\mathrm{ordered}$
   
   \UNTIL{Convergence}
\end{algorithmic}
\end{algorithm}
\begin{algorithm}
   \caption{The alternating conditioned iterative closest points for distribution training algorithm}
   \label{alg:alternating_incremental_matching}
\begin{algorithmic}[1]
   \REPEAT   
   \STATE Sample set $\mathcal{D}_\mathrm{batch\_origin} = \{\bx_0,\bx_1,...,\bx_N\}$ from $\mathcal{D}_\mathrm{origin}$
   \STATE Sample set \\$\mathcal{D}_\mathrm{batch\_target} = \left\{\begin{bmatrix}\bz_0 \\ \by_0\end{bmatrix},\begin{bmatrix}\bz_1 \\ \by_1\end{bmatrix},...,\begin{bmatrix}\bz_N \\ \by_N\end{bmatrix} \right\}$ \\from $\mathcal{D}_\mathrm{target}$
   \STATE Initialize $\mathcal{D}_\mathrm{pairs} = \emptyset$
   
   \FOR{$i = 0 ... N$}
		\STATE Perform the mapping $\begin{bmatrix}\widehat{\bz}_i \\ \byhat_i\end{bmatrix} = f\left( \begin{bmatrix}\bz_i \\ \bx_i \end{bmatrix} \right)$
		\STATE Add pair $\left(\begin{bmatrix}\bz_i \\ \bx_i \end{bmatrix},\begin{bmatrix}\bz_i \\ \byhat_i\end{bmatrix} \right)$ to $\mathcal{D}_\mathrm{pairs}$
   \ENDFOR
   
   \STATE Initialize $\mathcal{D}_\mathrm{ordered} = \emptyset$

   \WHILE{$|\mathcal{D}_\mathrm{batch\_target}| > 0$}
        \STATE Sample random number $b \sim \mathrm{Bernoulli}(0.5)$
        
     \IF { $b = 1$}
        	\STATE Pick random $\begin{bmatrix}\bz_i \\ \by_i\end{bmatrix} \in \mathcal{D}_\mathrm{batch\_target}$
        	\STATE Find index $\displaystyle j = \arg \left[ \min d \left( \begin{bmatrix}\bz_i \\ \by_i\end{bmatrix}, \begin{bmatrix}\bz_j \\ \byhat_j\end{bmatrix} \right) \right] $ of the set $\mathcal{D}_\mathrm{pairs}$
        \ELSE
        	\STATE Pick random $\left(\begin{bmatrix}\bz_j \\ \bx_j \end{bmatrix},\begin{bmatrix}\bz_j \\ \byhat_j \end{bmatrix} \right) \in \mathcal{D}_\mathrm{pairs}$
        	\STATE Find index $\displaystyle i = \arg \left[ \min d \left( \begin{bmatrix}\bz_i \\ \by_i\end{bmatrix}, \begin{bmatrix}\bz_j \\ \byhat_j\end{bmatrix} \right) \right] $ of the set $\mathcal{D}_\mathrm{batch\_target}$
        \ENDIF
        
        \STATE Add pair $\left(\begin{bmatrix}\bz_i \\ \bx_j \end{bmatrix},\begin{bmatrix}\bz_i \\ \by_i\end{bmatrix} \right)$ to $\mathcal{D}_\mathrm{ordered}$
       	\STATE Remove pair $\left(\begin{bmatrix}\bz_j \\ \bx_j \end{bmatrix},\begin{bmatrix}\bz_j \\ \byhat_j \end{bmatrix} \right)$ from $\mathcal{D}_\mathrm{pairs}$
   		\STATE Remove $ \begin{bmatrix}\bz_i \\ \by_i\end{bmatrix}$ from $\mathcal{D}_\mathrm{batch\_target}$
   \ENDWHILE   

   \STATE Train $f$ with supervised learning using $\mathcal{D}_\mathrm{ordered}$
   
   \UNTIL{Convergence}
\end{algorithmic}
\end{algorithm}

\FloatBarrier


\section{Results}
\label{sec:results}

This section presents some results obtained with the algorithm. All of the results were produced with a fully connected neural network with three hidden layers as the mapping $f$. The units of the hidden layers were chosen to be bipolar \citep{Eidnes2017} scaled exponential linear units \citep{Klambauer2017}. The neural neural net in Section \ref{sec:mnist} had 300 units per hidden layer. Otherwise the networks had 50 units per hidden layer. The supervised training was performed by ADAM algorithm \citep{Kingma2015}. We furthermore clipped the gradients of the output by clamping each dimension between -0.1 and 0.1 before backpropagation.

To highlight the flexibility of our the algorithm, we used a mixed value distribution as the origin distribution $\mathcal{D}_\mathrm{origin}$ in all our tests. Apart from Figures \ref{fig:teaser} and \ref{fig:mixed_input_illustration}, of the $N$ dimensions the first $N/2$ dimensions were sampled from a Bernoulli distribution with $p = 0.5$ and the last $N/2$ dimensions were sampled from the continuous uniform distribution. In Figures \ref{fig:teaser} and \ref{fig:mixed_input_illustration} the noise dimension $N$ is obviously 2, for the simple synthetic examples of Section \ref{sec:low_dim_examples} $N$ was 6, and otherwise it was 20.

\subsection{Low-dimensional examples}
\label{sec:low_dim_examples}

\begin{figure*}[h]
        \centering
        \begin{subfigure}{0.32\linewidth}
                \centering
                \includegraphics[width=\textwidth]{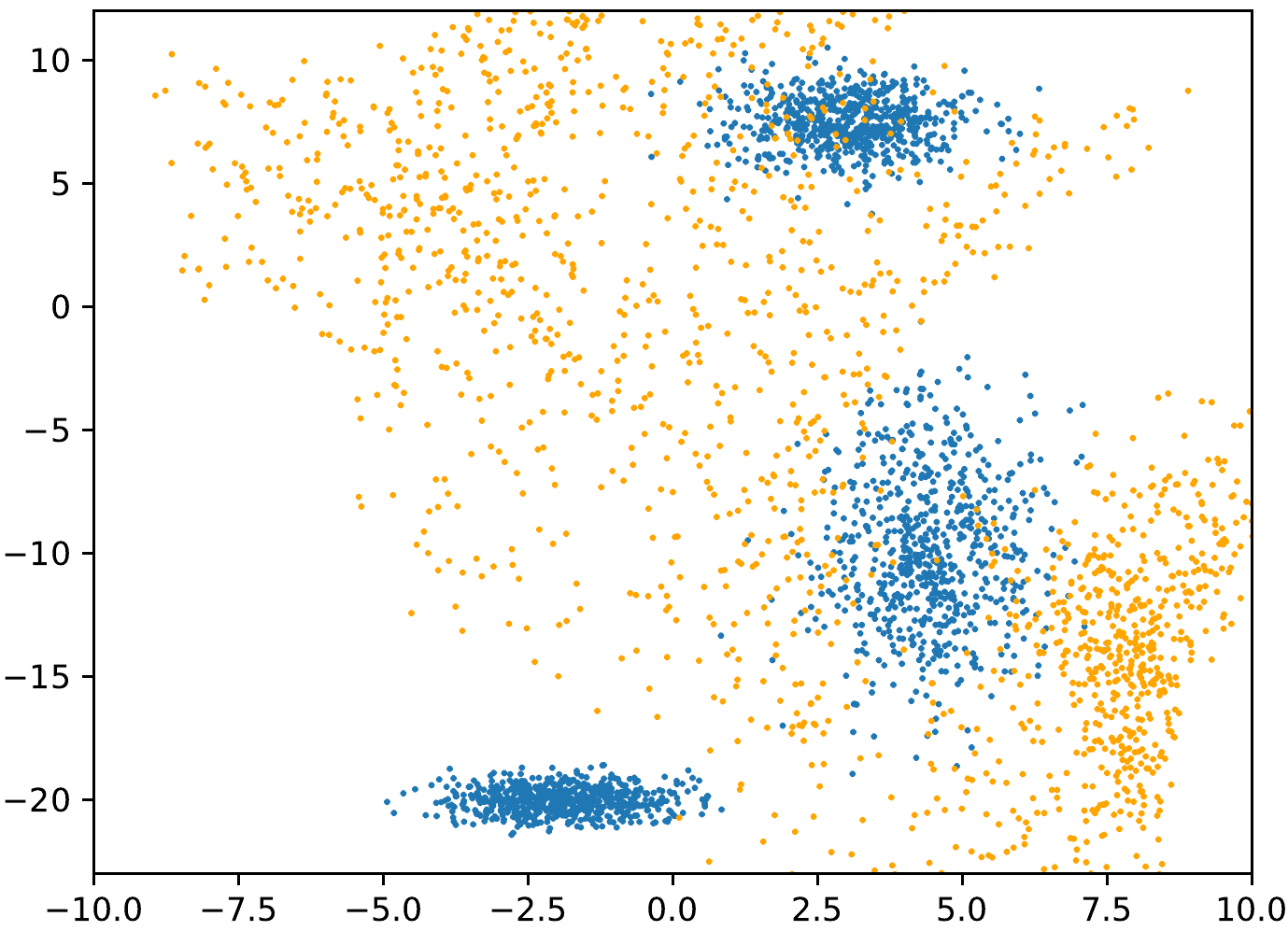}
                \caption{Initial situation.}
        \end{subfigure}%
        \begin{subfigure}{0.32\linewidth}
                \centering
                \includegraphics[width=\textwidth]{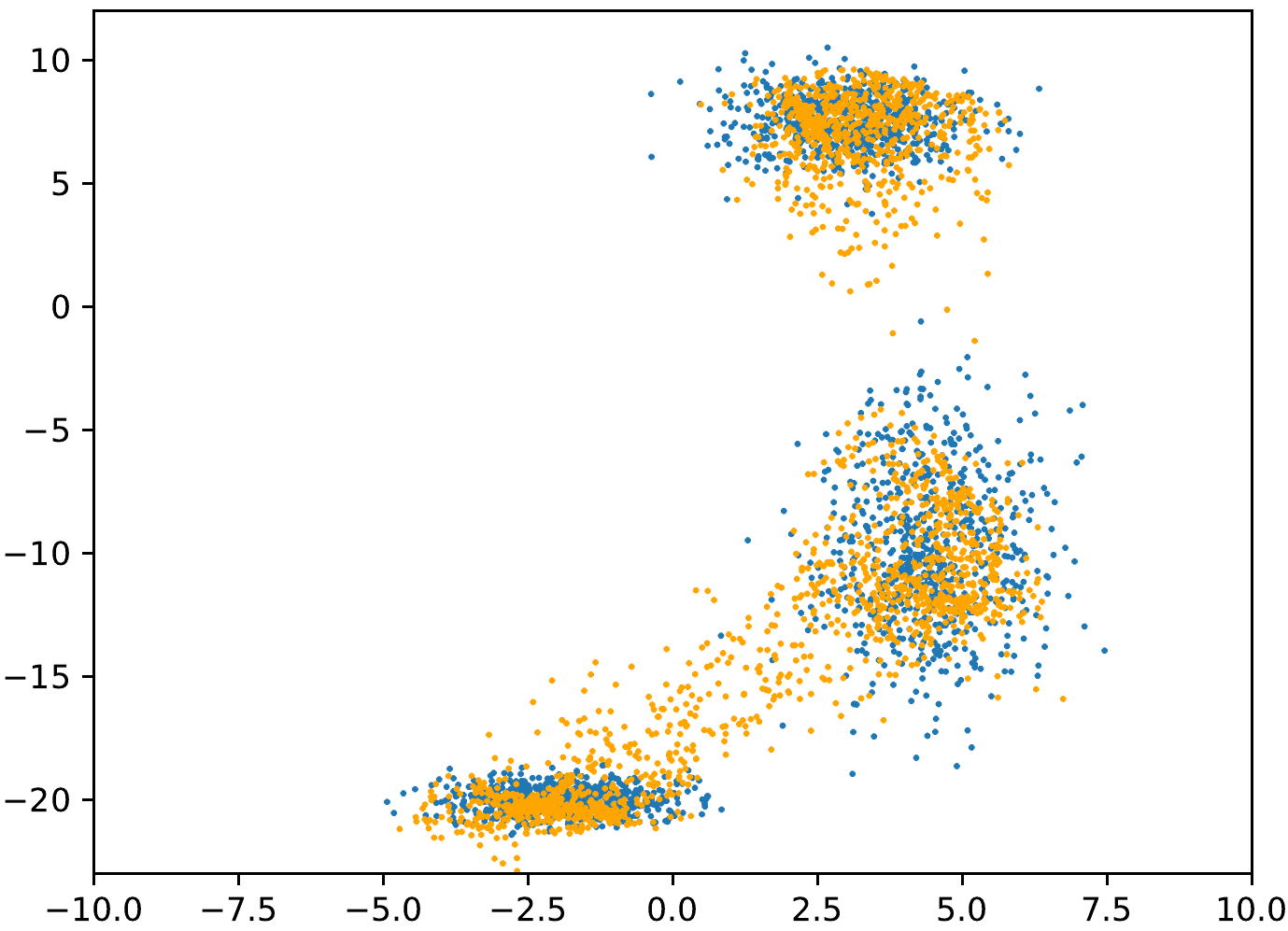}
                \caption{Epoch 5.}
        \end{subfigure}%
        \begin{subfigure}{0.32\linewidth}
                \centering
                \includegraphics[width=\textwidth]{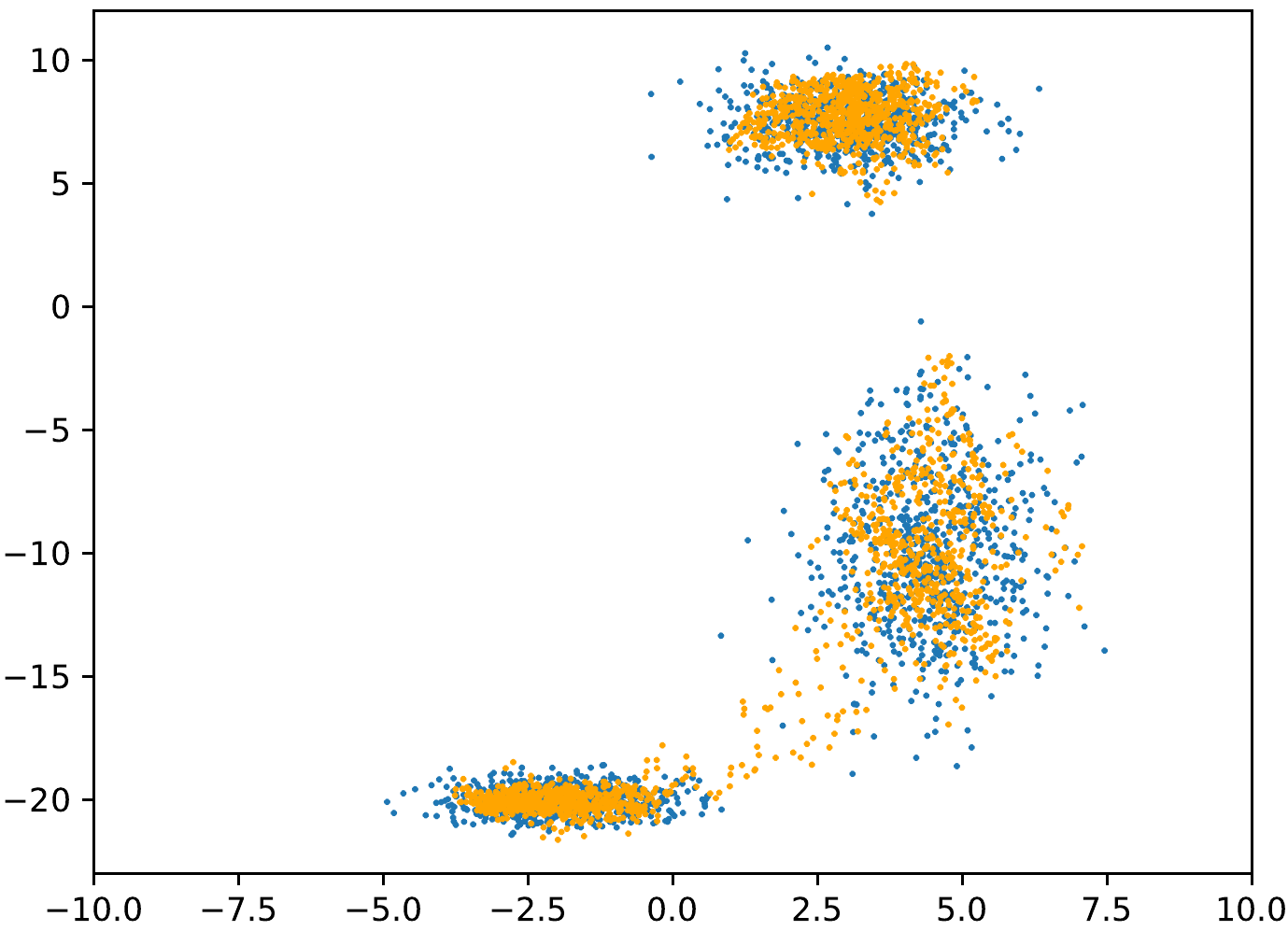}
                \caption{Epoch 50.}
        \end{subfigure}%
        ~ 
        \caption{A simple data set of three Gaussian distributions being approximated. The initial distribution has a very small overlap with the data, especially with one of the Gaussians. The algorithm improves fast and already after 5 epochs the sampled data starts to assume the form of the true data.\label{fig:three_gaussians}}
\end{figure*}

\begin{figure}
        \centering
        \includegraphics[width=0.5\textwidth]{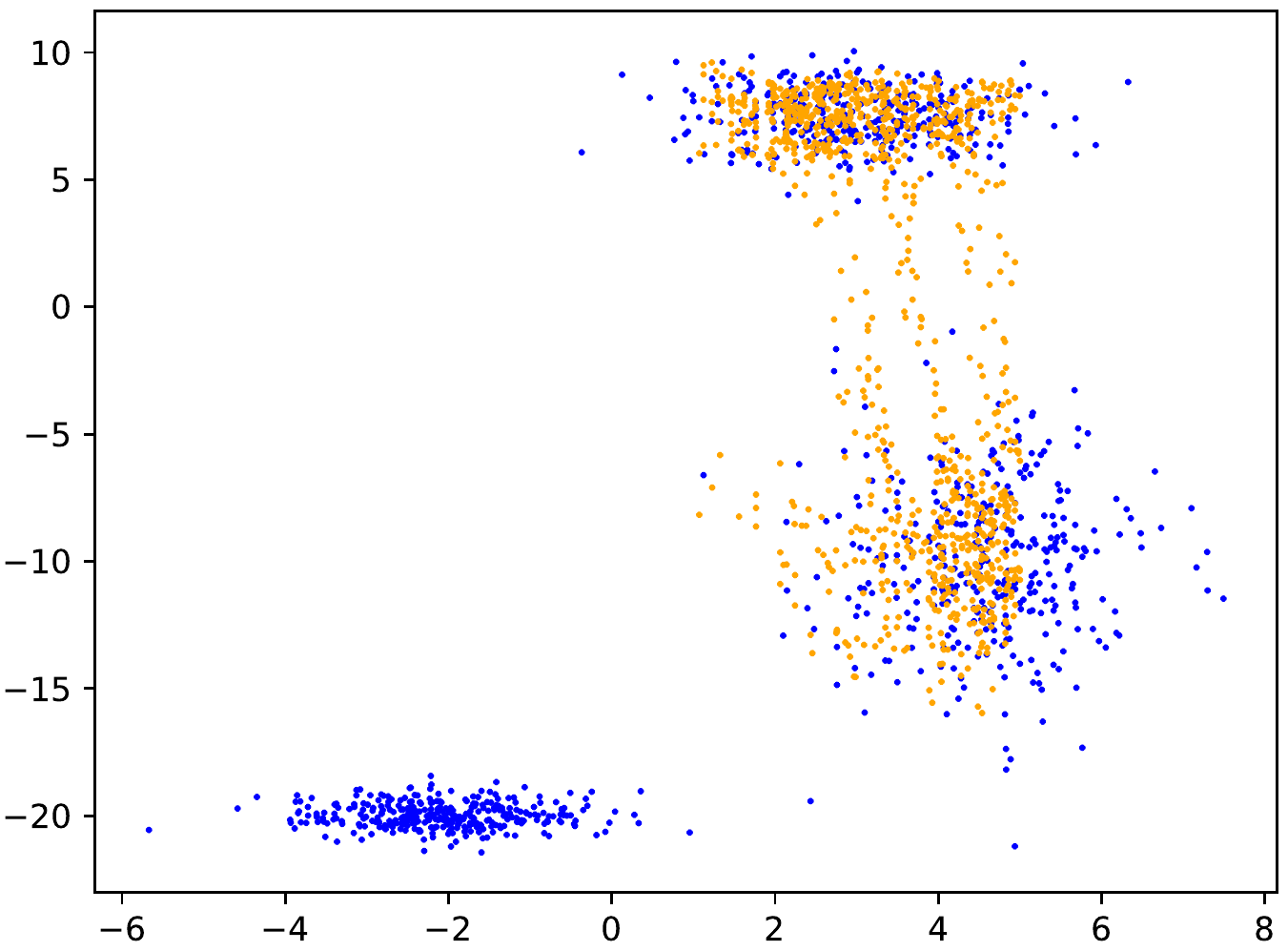}

        \caption{Conditioned data generation being demonstrated with a simple data set of three Gaussians. The x-axis value is selected from the interval [1,5] and the algorithm is asked to predict the missing y-axis value. \textcolor{blue}{Blue} denotes the ground truth data and \textcolor{cadmiumorange}{orange} is generated data. Some stray data points remain between the Gaussians but generally the algorithm captures the shape of the distribution well.\label{fig:three_gaussians_conditioned}}
\end{figure}

Figure \ref{fig:three_gaussians} presents a three component Gaussian mixture model and Figure \ref{fig:three_gaussians_conditioned} shows the same model with the model being conditioned by the x-axis variable. The distance metric used here was the Euclidean distance.

\begin{figure}
        \centering
        \includegraphics[width=0.5\textwidth]{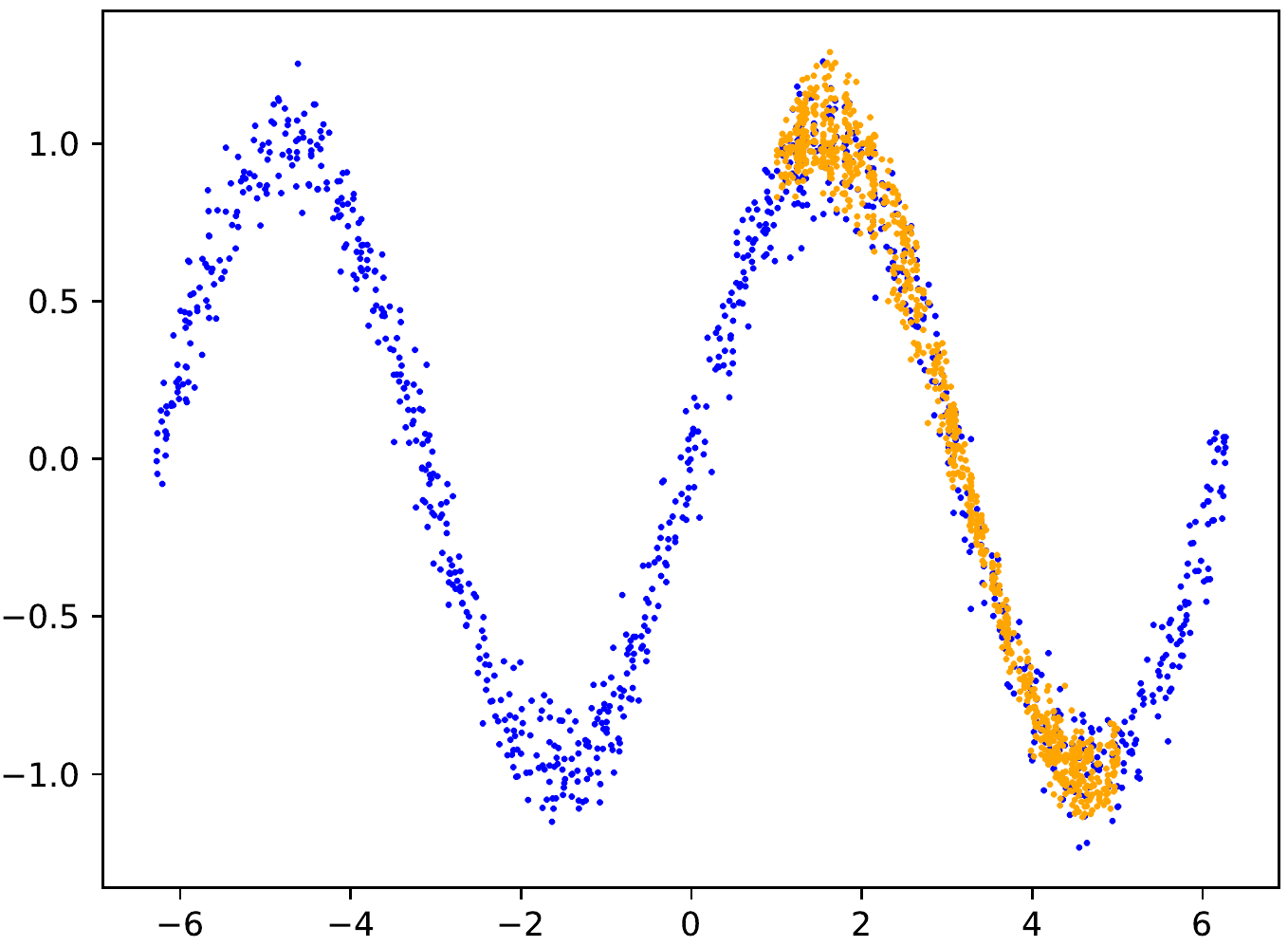}

        \caption{Demonstrating conditioned data generation with a sinusoid that has Gaussian noise added in the y-axis direction. The x-axis value is selected from the interval [1,5] and the algorithm is asked to predict the missing y-axis value. \textcolor{blue}{Blue} denotes the ground truth data and \textcolor{cadmiumorange}{orange} is generated data.\label{fig:serpent}}
\end{figure}

\begin{figure}
        \centering
        \begin{subfigure}{0.45\linewidth}
                \centering
                \includegraphics[width=\textwidth]{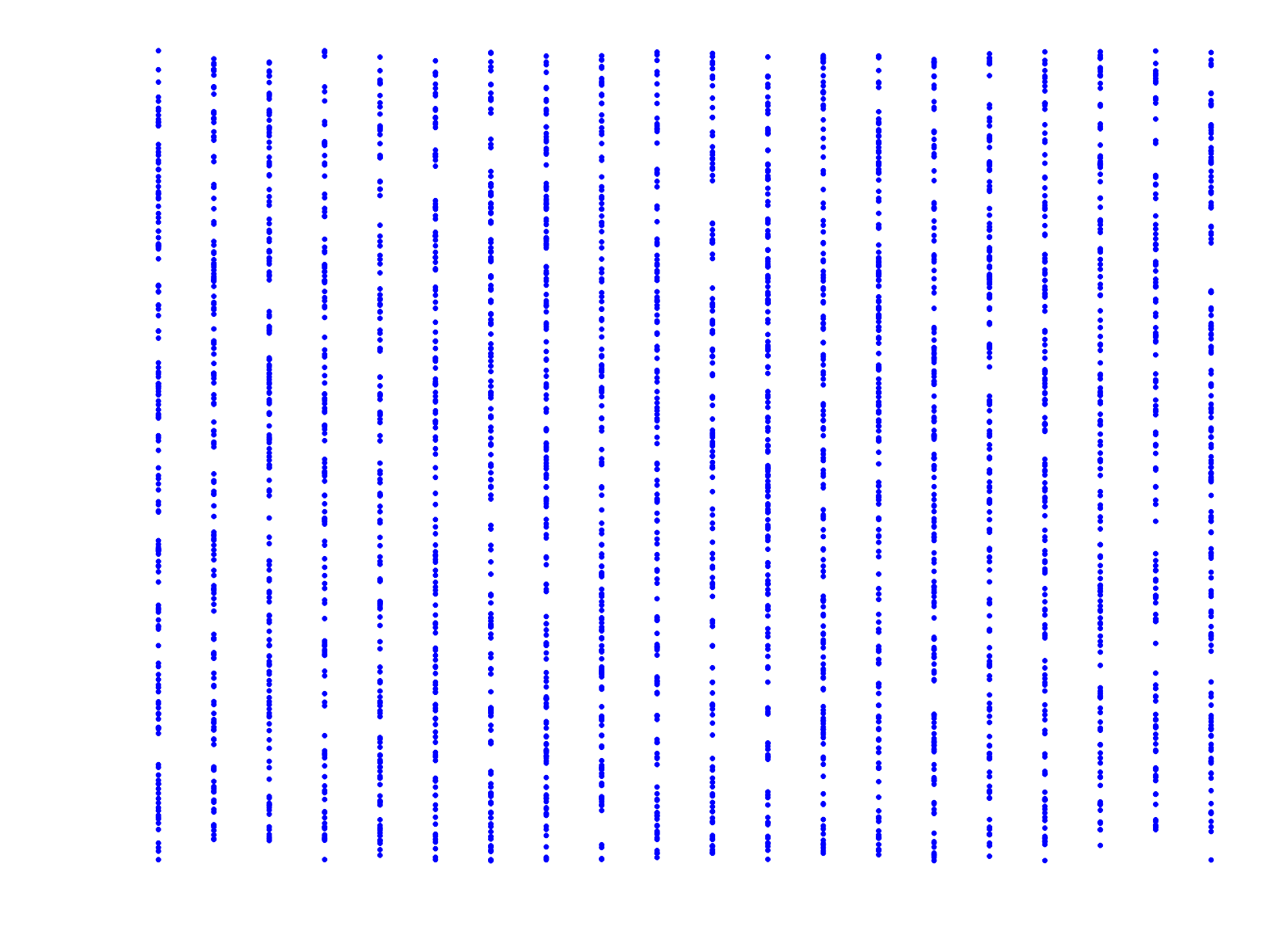}
                \caption{Samples from the origin distribution.}
        \end{subfigure}%
        {\LARGE$\Rightarrow$}
        \begin{subfigure}{0.45\linewidth}
                \centering
                \includegraphics[width=\textwidth]{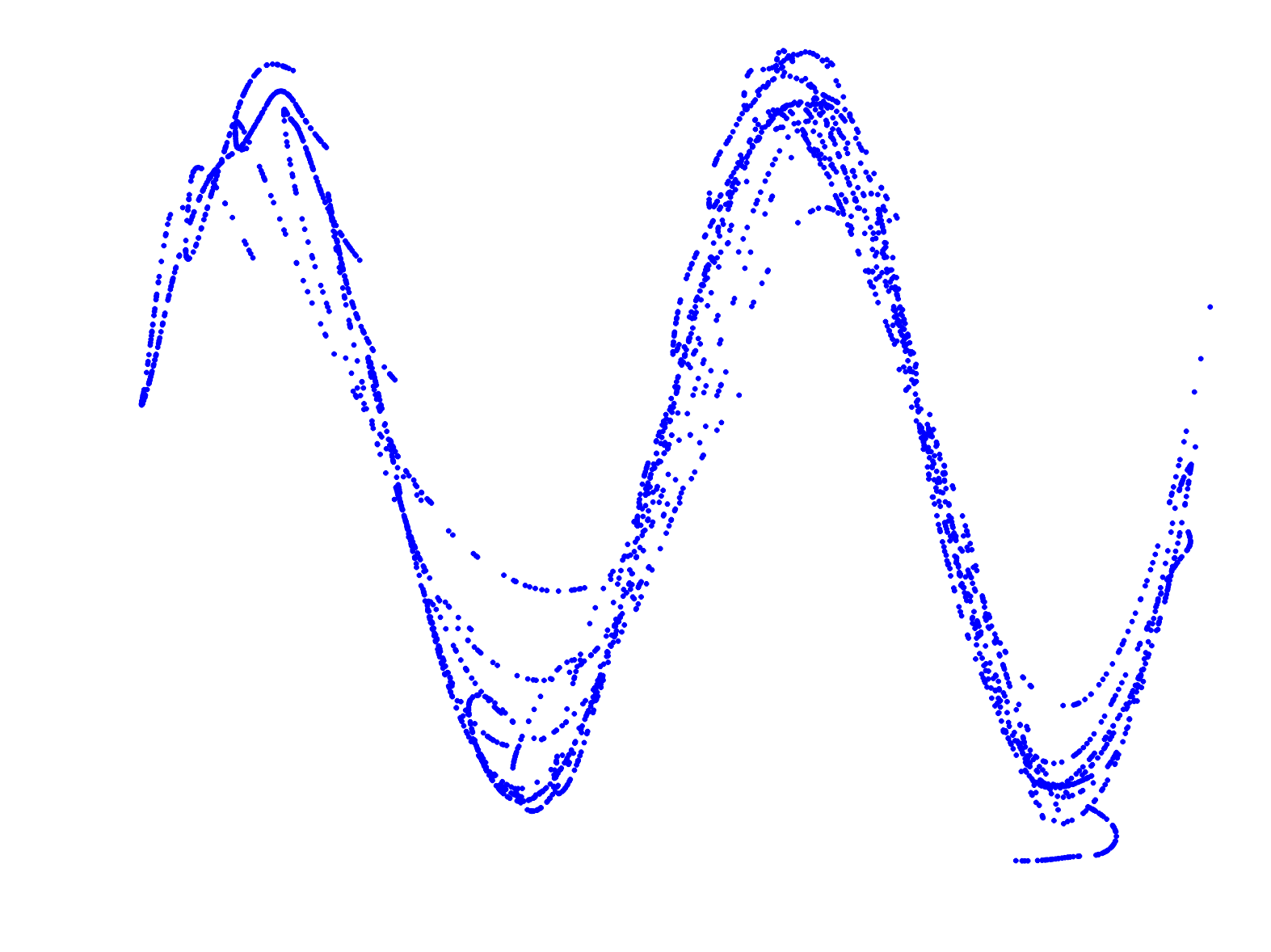}
                \caption{Generated samples.}
        \end{subfigure}%
        ~ 
        \caption{A two dimensional mixed continuous and discrete distribution is transformed to a sinusoid with added Gaussian noise. The stripy pattern of the input distribution translates also to the output distribution. \label{fig:mixed_input_illustration}}
\end{figure}

These examples illustrate how the algorithm is able to map a distribution lacking modalities to a three-modal distribution. Some stray data points remain between the distribution components because of the continuous mapping $f$. Figure shows \ref{fig:serpent} a sinusoid with added Gaussian noise. Here we observe no problems with the algorithm, as the target distribution does not exhibit multi-modality that would be in contrast to the input distribution. Figure \ref{fig:mixed_input_illustration} shows the same sinusoid based distribution being approximated with an adversarially chosen two dimensional mixed distribution with a discrete and a continuous variable. This figure also makes evident how the points that are close in the input space map close to each other in the output space.

In these low-dimensional examples we used a batch size of 500 in the training. The supervised training was done with minibatch size 100.


\FloatBarrier

\subsection{MNIST data}
\label{sec:mnist}

\begin{figure}
        \centering
   		\includegraphics[width=0.5\textwidth]{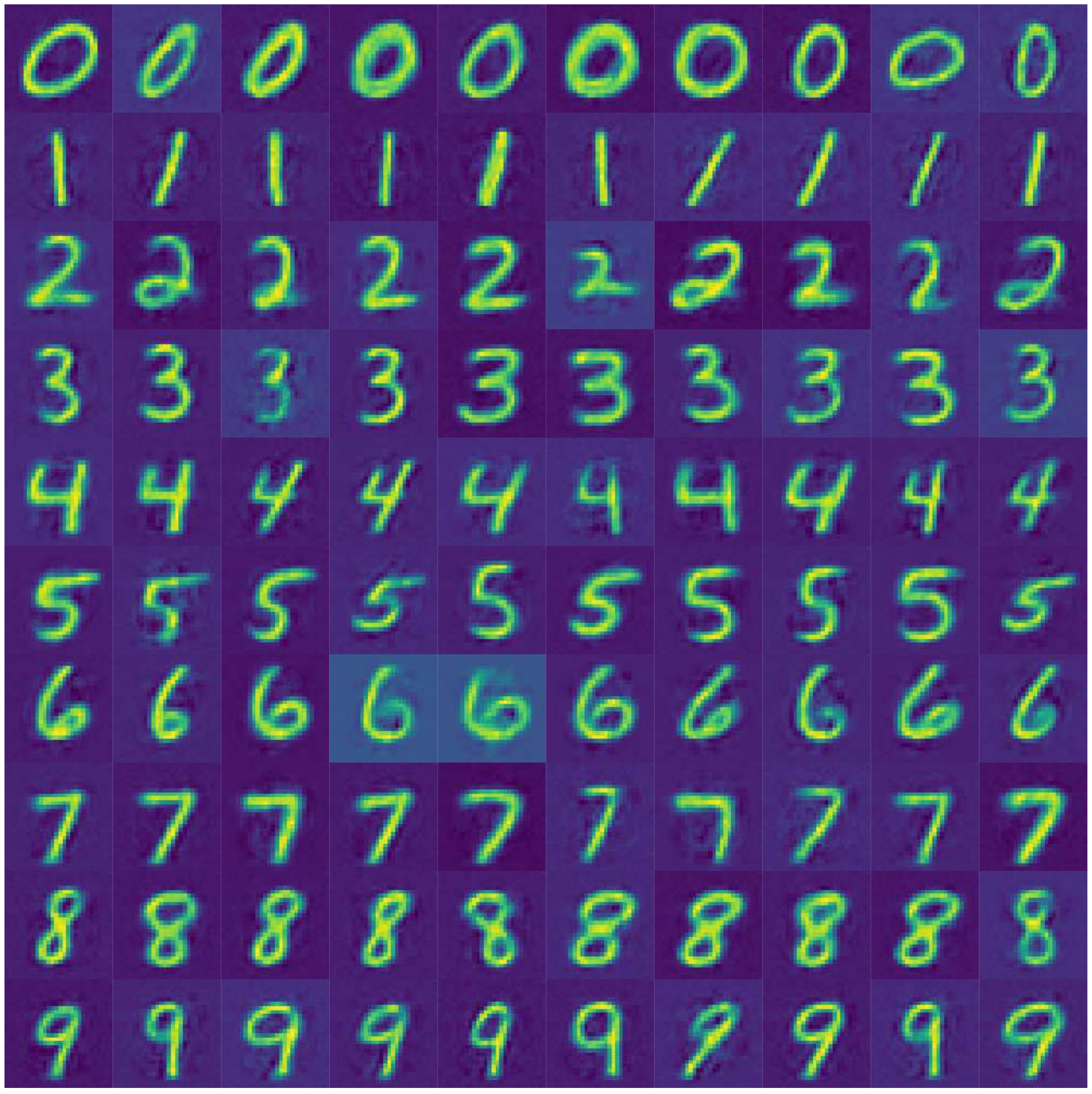}
        ~ 
        \caption{Images of the MNIST data. All of the images are produced by one densely connected neural network that has three hidden layers with 300 neurons each. We can generate images conditionally and here the network was asked to produce ten of each digit. We observe that the network can produce different characters and the variation within the character class. These figures are the results of epoch 250.\label{fig:mnist}}
\end{figure}

We also applied the method to the MNIST data \citep{mnist}. Some characters obtained this way are presented in Figure \ref{fig:mnist}. Also surprisingly to us, a simple fully connected neural network can produce all the characters with variation in each character.

The characters in Figure \ref{fig:mnist} have been generated conditionally, all by the same neural network. In this example we also used the Euclidean distance:
\begin{equation}
d\left(\begin{bmatrix} z_1 \\ \by_1 \end{bmatrix}, \begin{bmatrix}z_2 \\ \by_2 \end{bmatrix} \right) = (z_1 - z_2)^2 + (\by_1- \by_2)^\mathrm{T}(\by_1- \by_2).
\end{equation}
The Euclidean distance is not particularly good for image data. This is why it was surprising to see the algorithm perform quite well with a network architecture (fully connected) and a distance measure that are neither particularly good for the task at hand. 

\begin{figure}
        \centering
        \begin{subfigure}{0.4\linewidth}
                \centering
                \includegraphics[width=\textwidth]{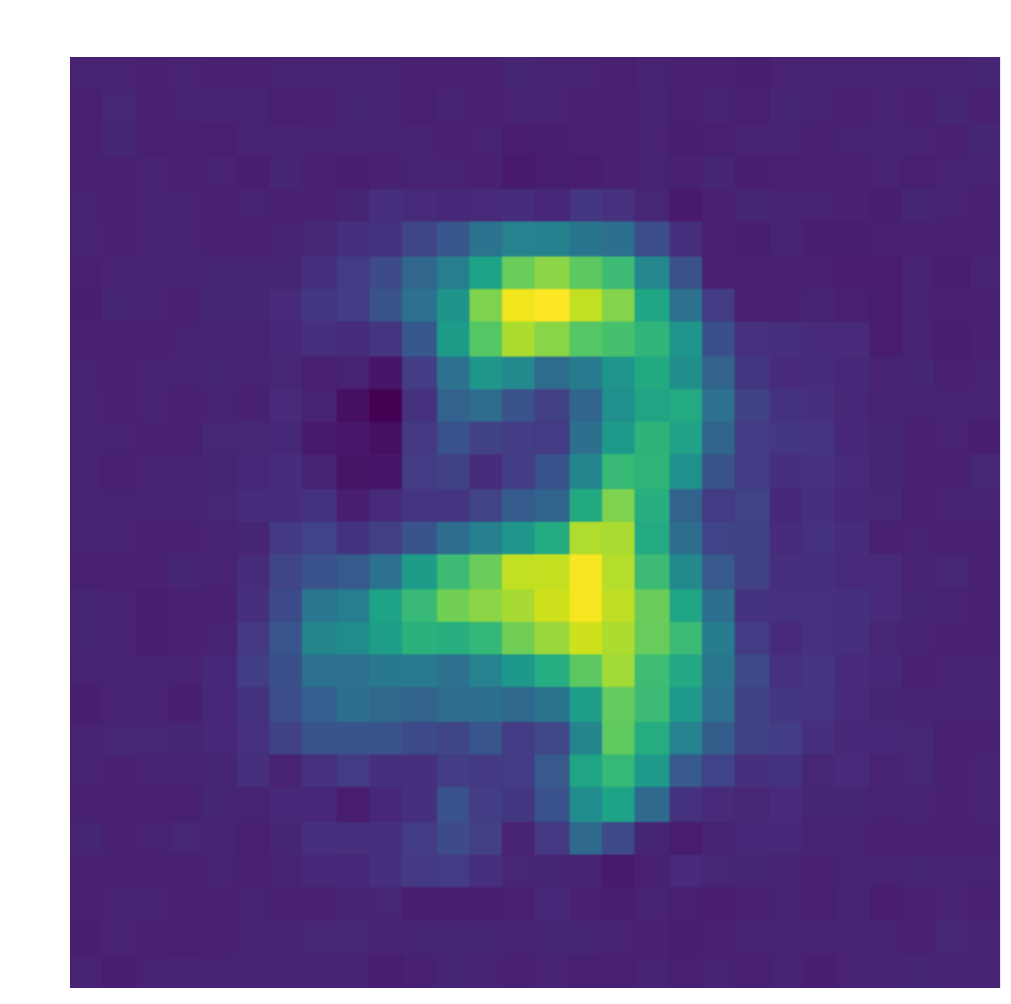}
                \caption{A character between 2 and 3. The character is supposed to be 2.}
        \end{subfigure}%
        \; \;
        \begin{subfigure}{0.4\linewidth}
                \centering
                \includegraphics[width=\textwidth]{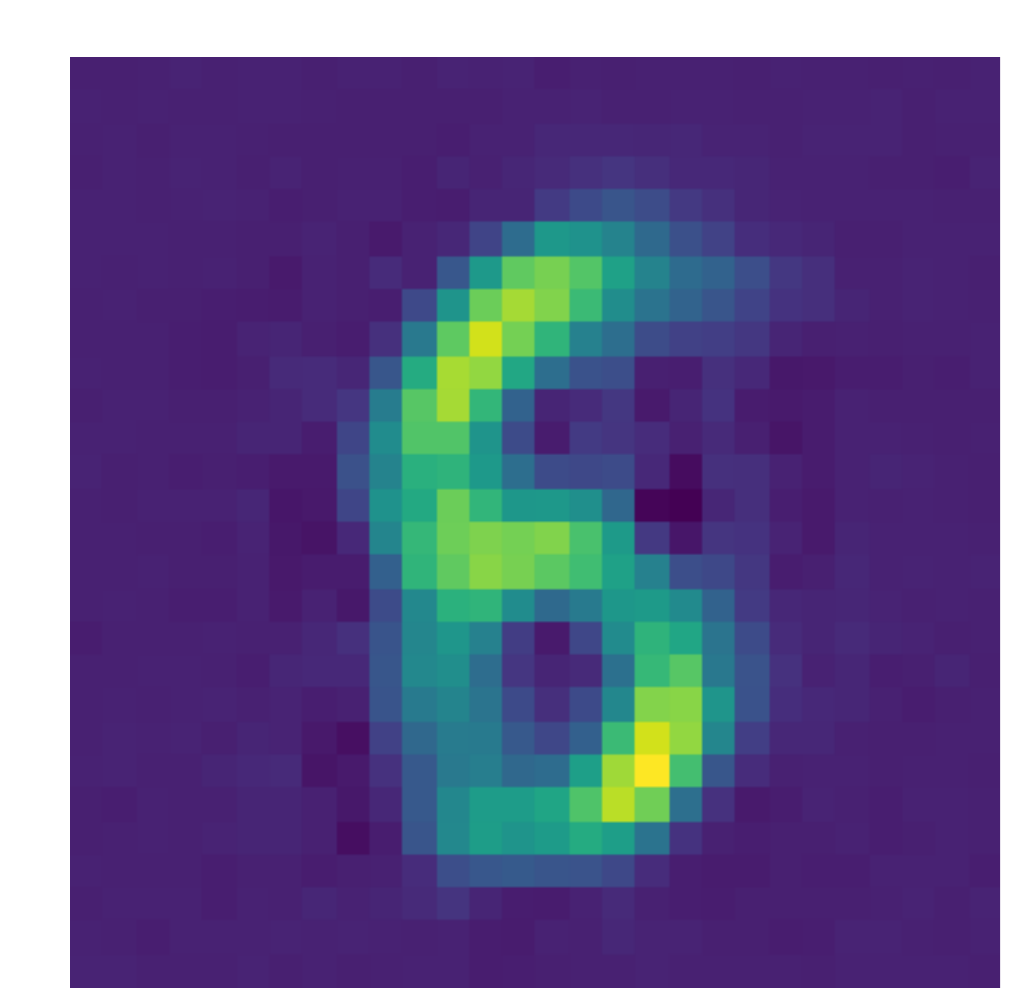}
                \caption{A character between 5 and 6. The character is supposed to be 6.}
        \end{subfigure}%
       
        ~ 
        \caption{In the early phase of the training (12th epoch) the neural net produces characters that seem to be mixes between two different categories. As the training progresses these mixes of two classes disappear.\label{fig:mnist_early}}
\end{figure}

The neural net starts to get some characters right already after the first epoch. Especially the characters that are distinctive from the others, such as zeros find their form early. In the early phase of the training the algorithm seems to produce characters that are mixtures of two different characters. A few examples are shown in Figure \ref{fig:mnist_early}. As the training progresses, these mixed forms become increasingly rare.

\begin{figure}
        \centering
   		\includegraphics[width=0.5\textwidth]{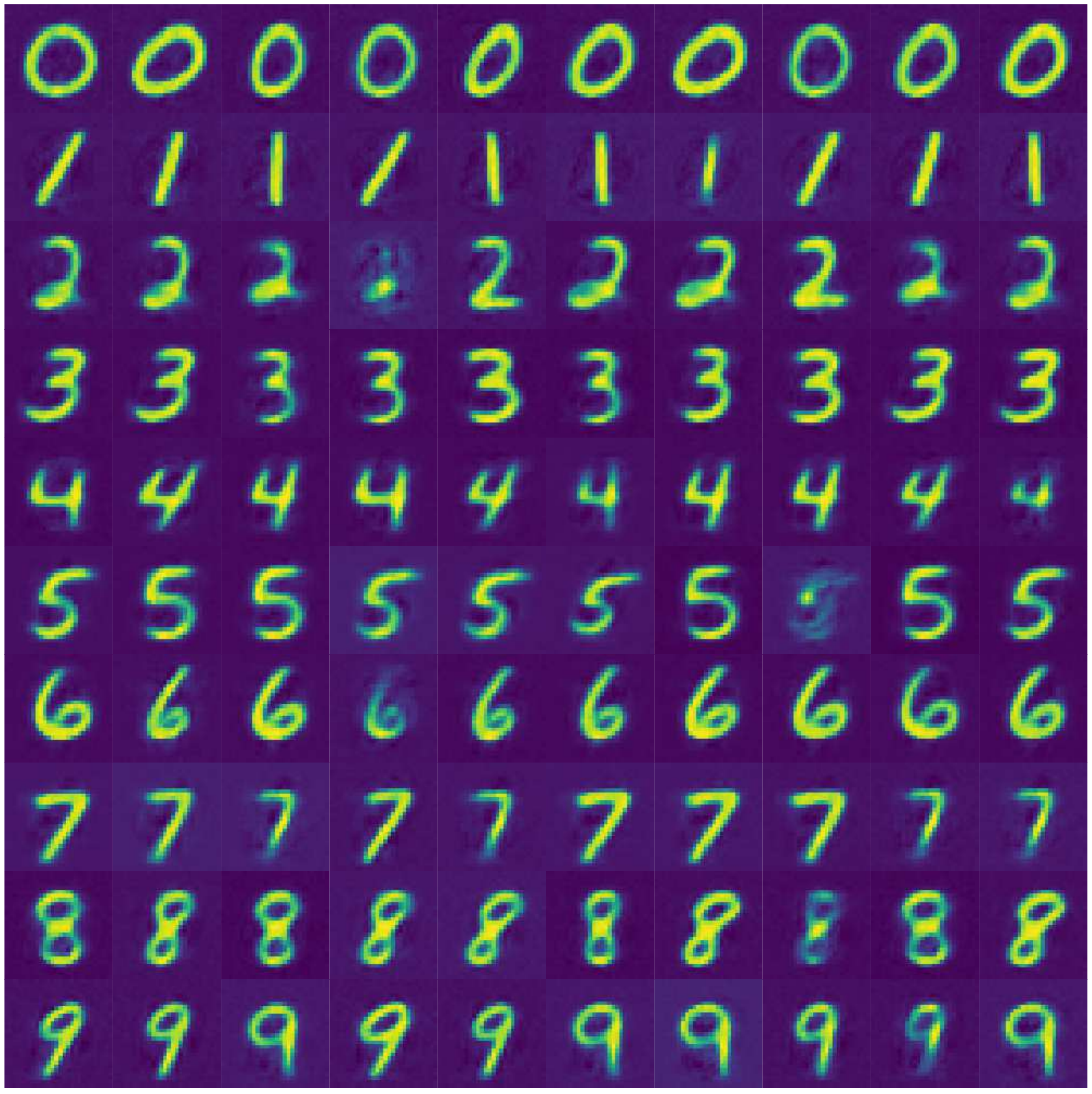}
        ~ 
        \caption{Images of the MNIST data. These figures have been produced with the correspondence finding batch size of 100 and the supervised learning batch size 100. The algorithm learns some variation in the data but to a lesser extent than with a bigger batch size.\label{fig:mnist_small_batch}}
\end{figure}

The training set of MNIST data contains 60,000 data points. We used the batch size 10,000 in the matching algorithm and the supervised training was performed with the minibatch size of 100. We also tried using a small minibatch size of 100 both in the matching algorithm and in the supervised learning, i.e. a simplified version of the algorithm where the correspondence finding batch and supervised training minibatch are the same and only one gradient step is taken for a set of corresponding sample pairs. The results are shown in Figure \ref{fig:mnist_small_batch}. The algorithm works to some extent but the results have smaller variation and the neural network produces some quite faint characters.


\FloatBarrier

\subsection{Learning categorical distributions}
\label{sec:continuous_to_discrete}

\begin{figure*}[h]

        \centering
        \begin{subfigure}{0.45\linewidth}
                \centering
                \includegraphics[width=\textwidth]{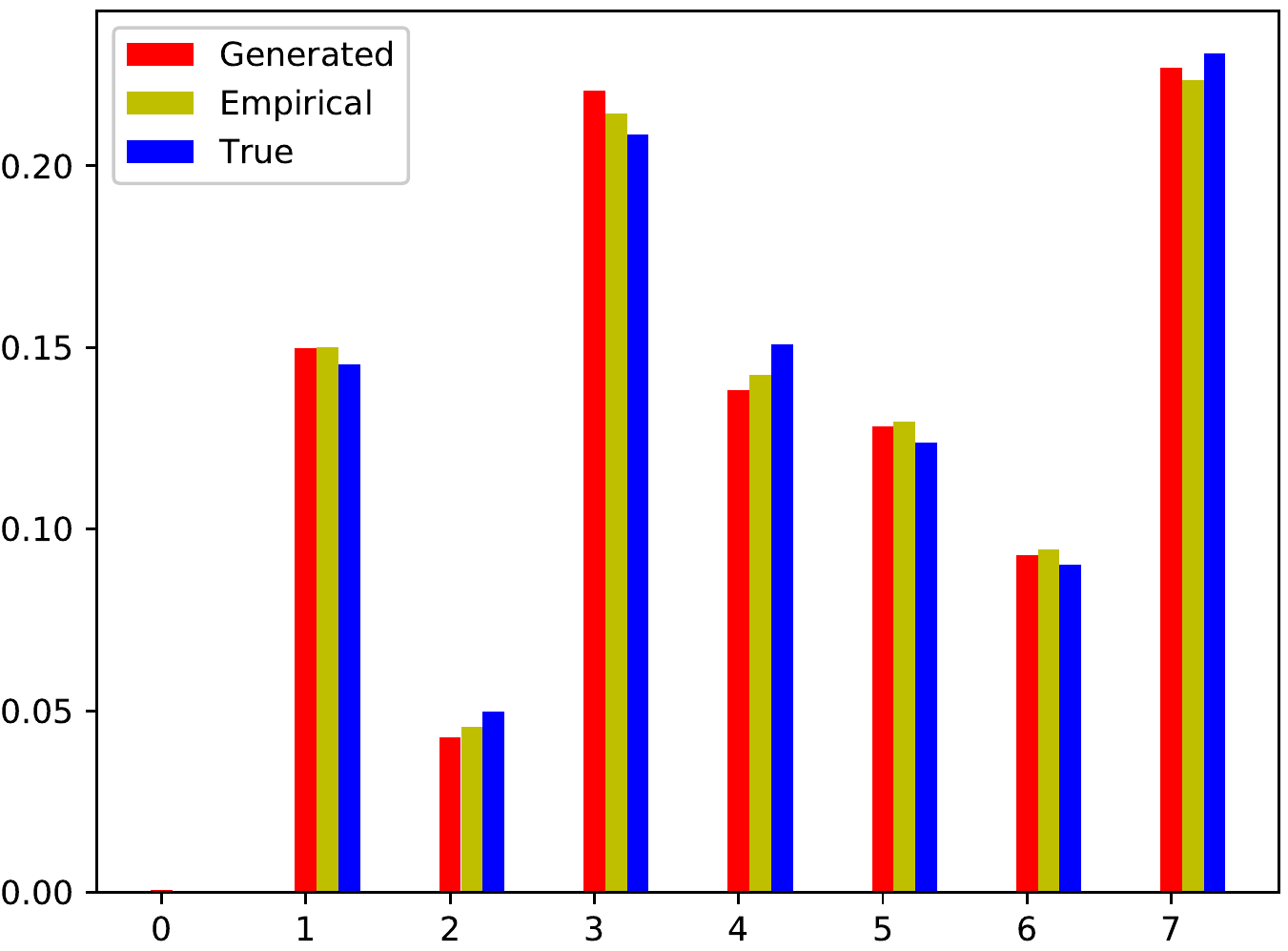}
                \caption{A 1000 data point sample generated by our algorithm and a 1000 data point empirical sample that has been sampled from the ground truth distribution. The plot is also showing the probabilities of the ground truth distribution. \label{fig:continuous_to_discrete_sample}}
        \end{subfigure}%
        \qquad
        \begin{subfigure}{0.45\linewidth}
                \centering
                \includegraphics[width=\textwidth]{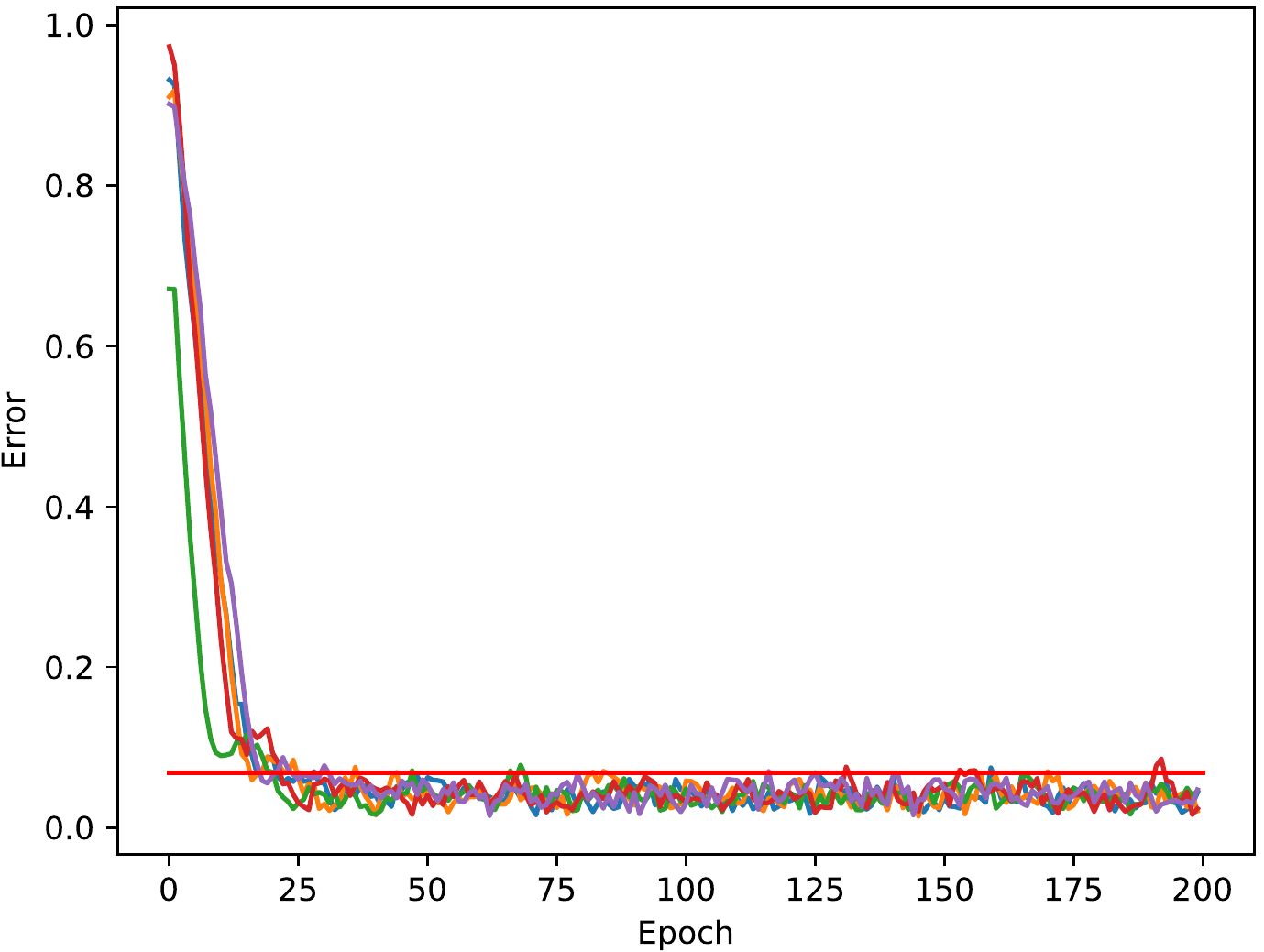}
                \caption{Convergence curves of a few runs on the algorithm. The horizontal red line indicates the mean of errors between the true pmf and a 1000-sample empirical pmf.\label{fig:continuous_to_discrete_convergence}}
        \end{subfigure}%
        
        ~ 
        \caption{Demonstrating learning a discrete distribution, in this case a Multinoulli distribution.\label{fig:continuous_to_discrete_demonstration}}
\end{figure*}

We also tested how our algorithm would fit to learn categorical distributions. Figure \ref{fig:continuous_to_discrete_sample} shows the result of one such distribution. Figure \ref{fig:continuous_to_discrete_convergence} shows a few convergence curves.

The distance measure that was used here was different from the cases seen previously. The neural network output was of the same dimension as the number of the categories in the distribution. We interpreted the maximum element of the prediction $\widehat{\by}$ to denote the selected class. The distance used here was the "cross-entropy" of softmax:
\begin{eqnarray}
d(\by,\widehat{\by}) = -\sum_i y_i \log \widetilde{y}_i\\
\widetilde{\by} = \text{softmax}(\widehat{\by}).
\end{eqnarray}
Here $\by$ is a one-hot vector indicating the class.


\FloatBarrier


\section{Discussion}
\label{sec:discussions}

We have presented a simple algorithm to train a continuous mapping from a nearly arbitrary sampling distribution to a target distribution. The algorithm works for cases with a moderate amount of dimensions. It also necessitates that there is a meaningful distance for the distribution which we wish to approximate. This is not the case for example for natural images but there are for instance reinforcement learning algorithms that could benefit from our algorithm \citep{Rajamaki2017}.

Because we are using supervised learning to train the mapping $f$, the algorithm is very robust, assuming that point-to-point correspondences are consistent between iterations, similarly to the stochastic minibatches in supervised training. Furthermore, because each data point is matched to a point in the input distribution, the algorithm cannot ignore any parts of the target distribution. Our simple synthetic example in Figure \ref{fig:three_gaussians} demonstrates how the algorithm converges even though there is minuscule overlap between the initially generated data and the true data.

Our algorithm has a side product that is of interest in itself. The algorithms presented in this paper produce a matching of the mapped points and the true data points. We can compute the sum of the distances between the true data points and the corresponding mapped data points. This serves as a quantity measuring the convergence of the algorithm. Thus, we have a quantitative way of measuring the convergence.

\subsection{Limitations}

Each batch of data selected for the training procedure should be representative of the whole distribution being approximated. Likewise each batch of noise should be representative of the whole origin distribution. As seen in Section \ref{sec:mnist}, the algorithm works with small minibatches, but as is the case with supervised learning, the quality is improved with a bigger batch size. The minibatch size of the supervised learning part of the algorithm can be chosen independently of the correspondence finding algorithm's batch size, though. Nonetheless, because of the quadratic computational cost of the correspondence finding algorithm, the balancing between a batch size that captures the variation in the data and a batch size that yields fast training becomes even more pronounced than with pure supervised learning.

The algorithm necessitates the existence of a meaningful distance in the space of the distribution being approximated. This is not the case for high-dimensional distributions such as distributions consisting of high-resolution images. Our algorithm works well with a moderate amount of dimensions, though, as demonstrated by the example with the MNIST data.

\section{Conclusions}
\label{sec:conclusions}

We have presented a robust and simple training algorithm to transform distributions in a moderate dimensional setting. The algorithm works well with and without conditioning variables and involves only one function approximator, such as a neural network. Be believe that there is still plenty of room for improvement in the algorithm, especially in developing correspondence procedures and distance metrics that produce consistent results with smaller minibatch sizes and high-dimensional data.

\FloatBarrier

\section*{Acknowledgements}



\vskip 0.2in
\bibliography{distribution_mapping_bibliography}

\end{document}